%% file: main.tex
\documentclass[nohyperref]{article}

\usepackage{microtype}
\usepackage{graphicx}
\usepackage{subfigure}
\usepackage{booktabs} 

\usepackage{hyperref}

\usepackage[accepted]{icml2022}

\usepackage{amsmath}
\usepackage{amssymb}
\usepackage{mathtools}
\usepackage{amsthm}
\usepackage{bm}
\usepackage{graphicx}
\usepackage{diagbox}
\usepackage{multirow}
\usepackage{enumitem}

\usepackage[capitalize,noabbrev]{cleveref}

\theoremstyle{plain}
\newtheorem{theorem}{Theorem}[section]
\newtheorem{proposition}[theorem]{Proposition}
\newtheorem{lemma}[theorem]{Lemma}
\newtheorem{corollary}[theorem]{Corollary}
\theoremstyle{definition}
\newtheorem{definition}[theorem]{Definition}
\newtheorem{assumption}[theorem]{Assumption}
\theoremstyle{remark}
\newtheorem{remark}[theorem]{Remark}

\input{math_commands}

\icmltitlerunning{Fighting Fire with Fire: Avoiding DNN Shortcuts through Priming}

\begin{document}

\twocolumn[
\icmltitle{Fighting Fire with Fire: Avoiding DNN Shortcuts through Priming}

\icmlsetsymbol{equal}{*}

\begin{icmlauthorlist}
\icmlauthor{Chuan Wen}{iiis}
\icmlauthor{Jianing Qian}{upenn}
\icmlauthor{Jierui Lin}{austin}
\icmlauthor{Jiaye Teng}{iiis}
\icmlauthor{Dinesh Jayaraman}{upenn}
\icmlauthor{Yang Gao}{iiis,qizhi}
\end{icmlauthorlist}

\icmlaffiliation{iiis}{Institute for Interdisciplinary Information Sciences, Tsinghua University}
\icmlaffiliation{upenn}{University of Pennsylvania}
\icmlaffiliation{austin}{University of Texas at Austin}
\icmlaffiliation{qizhi}{Shanghai Qi Zhi Institute}

\icmlcorrespondingauthor{Yang Gao}{gaoyangiiis@tsinghua.edu.cn}

\icmlkeywords{Machine Learning, ICML}

\vskip 0.3in
]



\printAffiliationsAndNotice{}  

\begin{abstract}
Across applications spanning supervised classification and sequential control, deep learning has been reported to find ``shortcut'' solutions that fail catastrophically under minor changes in the data distribution. In this paper, we show empirically that DNNs can be coaxed to avoid poor shortcuts by providing an additional ``priming'' feature computed from key input features, usually a coarse output estimate. Priming relies on approximate domain knowledge of these task-relevant key input features, which is often easy to obtain in practical settings. For example, one might prioritize recent frames over past frames in a video input for visual imitation learning, or salient foreground over background pixels for image classification. On NICO image classification, MuJoCo continuous control, and CARLA autonomous driving, our priming strategy works significantly better than several popular state-of-the-art approaches for feature selection and data augmentation. We connect these empirical findings to recent theoretical results on DNN optimization, and argue theoretically that priming distracts the optimizer away from poor shortcuts by creating better, simpler shortcuts.
Project website: \url{https://sites.google.com/view/icml22-fighting-fire-with-fire/}.
\end{abstract}

\input{1_introduction}

\input{3_method}

\input{4_experiments}

\input{2_related_work}

\input{5_conclusion}

\section{Acknowledgements}
We thank the anonymous reviewers for their helpful feedback.
This work is supported by the Ministry of Science and Technology of the People's Republic of China, the 2030 Innovation Megaprojects ``Program on New Generation Artificial Intelligence" (Grant No. 2021AAA0150000), and a grant from the Guoqiang Institute, Tsinghua University.

\bibliography{example_paper}
\bibliographystyle{icml2022}

\input{6_appendix}

\end{document}

%% file: math_commands.tex
\newcommand{\x}{\boldsymbol{x}}
\newcommand{\z}{\boldsymbol{z}}
\newcommand{\X}{\boldsymbol{X}}
\newcommand{\y}{y}
\newcommand{\Y}{\boldsymbol{Y}}
\newcommand{\Z}{\boldsymbol{Z}}
\newcommand{\W}{\boldsymbol{W}}
\newcommand{\w}{\boldsymbol{w}}
\newcommand{\vb}{\boldsymbol{v}}

\usepackage{hyperref}       
\usepackage{amsfonts}
\usepackage{amsmath}
\usepackage{amsthm}
\usepackage{bm}
\usepackage{float}
\usepackage{amssymb}
\usepackage{graphicx} 
\usepackage{nccmath}
\usepackage{thmtools}
\usepackage{thm-restate}
\usepackage{color}
\usepackage{cleveref}

\newcommand{\cN}{\mathcal{N}}

\newcommand{\cP}{\mathcal{P}}

\newcommand{\bE}{\mathbb{E}}

\newcommand{\bP}{\mathbb{P}}

\newcommand{\bR}{\mathbb{R}}

\usepackage{mathtools}
\usepackage{amssymb}
\usepackage{latexsym}
\usepackage{dsfont}
\usepackage{mathrsfs}
\usepackage{amssymb}
\usepackage{amsfonts}
\usepackage{bm}
\usepackage{xspace}

\usepackage{pifont}

        {\medskip}

        {\hspace*{\fill}$\Box$\par\vspace{4mm}}
        {\hspace*{\fill}$\Box$\par}

\ifx\BlackBox\undefined
\newcommand{\BlackBox}{\rule{1.5ex}{1.5ex}}  %
\fi

\ifx\QED\undefined
\def\QED{~\rule[-1pt]{5pt}{5pt}\par\medskip}
\fi

\ifx\theorem\undefined
\newtheorem{theorem}{Theorem}[section]
\fi

\ifx\example\undefined
\newtheorem{example}{Example}[section]
\fi

\ifx\property\undefined
\newtheorem{property}{Property}
\fi

\ifx\lemma\undefined
\newtheorem{lemma}{Lemma}[section]
\fi

\ifx\proposition\undefined
\newtheorem{proposition}{Proposition}[section]
\fi

\ifx\remark\undefined
\newtheorem{remark}{Remark}
\fi

\ifx\corollary\undefined
\newtheorem{corollary}{Corollary}[section]
\fi

\ifx\definition\undefined
\newtheorem{definition}{Definition}[section]
\fi

\ifx\conjecture\undefined
\newtheorem{conjecture}{Conjecture}[section]
\fi

\ifx\axiom\undefined
\newtheorem{axiom}[theorem]{Axiom}
\fi

\ifx\claim\undefined
\newtheorem{claim}[theorem]{Claim}
\fi

\ifx\assumption\undefined
\newtheorem{assumption}{Assumption}
\fi

\ifx\problem\undefined
\newtheorem{problem}{Problem}[section]
\fi

\ifx\fact\undefined
\newtheorem{fact}{Fact}
\fi

%% file: 1_introduction.tex
\section{Introduction}
Supervised deep neural networks (DNNs) have led to remarkable achievements across many applications~\citep{he2016deep,schrittwieser2020mastering,brown2020language}. Today's DNNs learn from a very limited form of supervision, namely, target annotations for each sample in a training dataset. In contrast, when humans learn from annotated examples, they often benefit from richer supervision. When teaching a child to recognize a ``zebra'' in a zoo enclosure, a parent might draw the child's attention to the animal and its black-and-white stripes. Without such priming, the learning task is much more poorly defined and there are many equally plausible concepts that the child might learn to identify as ``zebra'', such as a zoo enclosure, or any large equine animal including horses and donkeys.

\begin{figure}[t]
    \includegraphics[width=\linewidth]{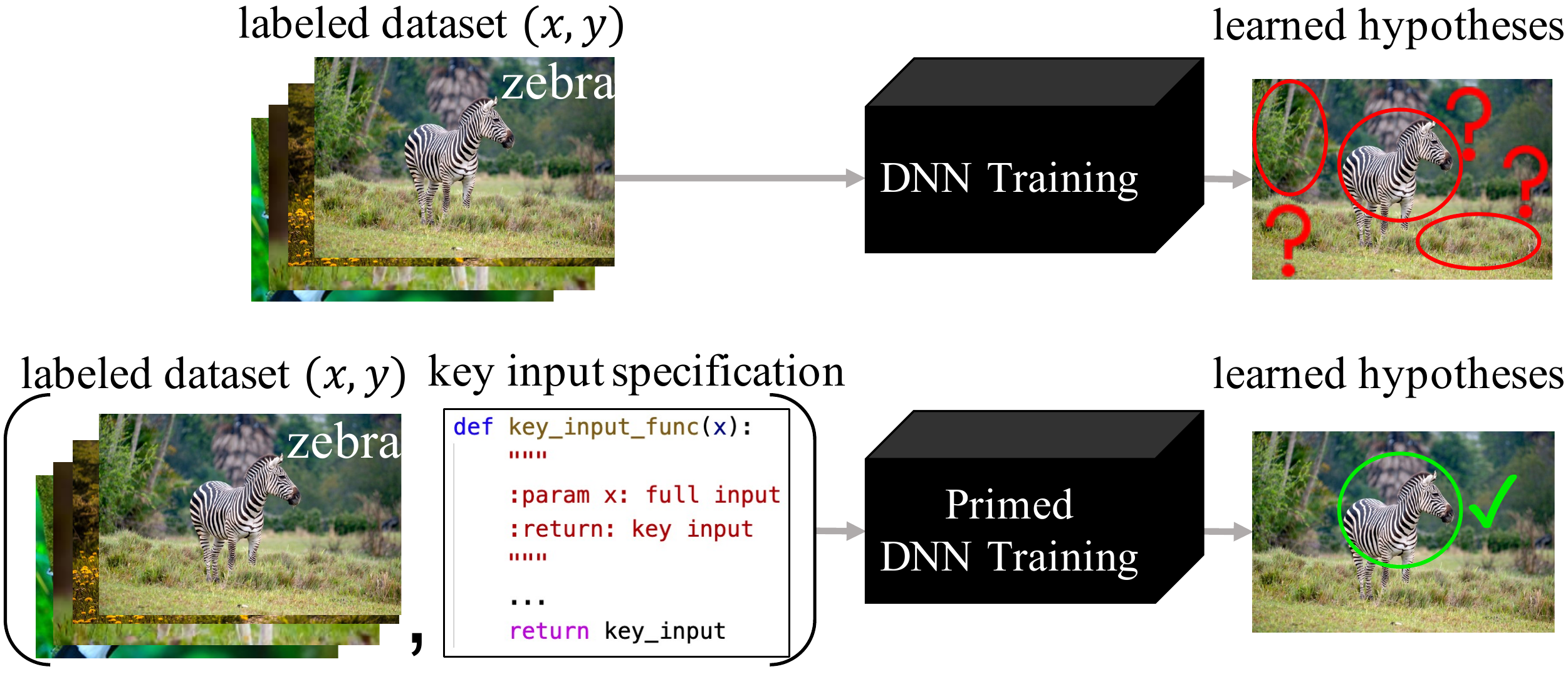}
  \caption{DNNs trained only from label annotation supervision often struggle to disambiguate between competing hypotheses for a target concept, and end up learning ``shortcuts'', such as classifying the background as the ``zebra''. 
  We propose to expand DNN supervision to include a key input specification function that identifies important portions of the input (such as the foreground object) according to domain knowledge,
  leading the optimization towards the intended solution.
  }
  \label{fig:concept}
\end{figure}

Indeed, modern DNNs suffer from problems very reminiscent of this confused child. For example, \citet{beery2018recognition} reports that DNNs trained for animal image recognition successfully classify cows on grass but not on beaches, because they wrongly attribute the label ``cow'' to the grass. Similarly, autonomous driving DNNs, rather than generating new steering actions in response to changing road observations, commonly learn to copy the previous action, since adjacent actions in training data were almost always very similar~\citep{chuan2020fighting, bansal2018chauffeurnet, codevilla2019exploring}. We review more such examples in Section~\ref{sec:preliminaries}. In all these cases, the annotated examples in the training dataset alone cannot sufficiently distinguish the correct solution and prevent the DNN from learning the wrong thing. Following~\citet{geirhos2020shortcut}, we use the term ``shortcut issue'' to refer to these phenomena where DNNs fail catastrophically under small distributional shift from training data. How to generalize beyond training distributions is a uniting fundamental question driving many decades-old sub-disciplines of machine learning research, including distributional robustness, domain generalization, zero-shot learning, causality, and invariant learning~\citep{Shen2021TowardsOG,Wang2021GeneralizingTU}.

In this work, we argue that many practically encountered ML shortcuts can be circumvented by a simple fix: richer supervision by providing auxiliary ``priming'' information, beyond merely the target labels. \textit{What form} should such priming information take? Much like the parent in the introductory example, we propose to exploit commonly available domain knowledge of where in the input the task-relevant information is most likely to lie, for example, the salient foreground for an image classification task, or the most recent observations in autonomous driving. While these ``key inputs'' do not typically contain \textit{all} relevant information, and may occasionally not even contain the most relevant information,\footnote{such as a pedestrian who is occluded behind a car in the most recent observation, but who was visible in an older observation} we find that this is not a problem. It is sufficient that they are likely in many cases to contain key information.

Having decided upon key input-based priming information, \textit{how} should we provide it to a target DNN learner to help avoid shortcuts? We propose to fight fire with fire by creating a new shortcut that biases the DNN optimization process towards solutions that pay attention to the key inputs. In practice, our ``PrimeNet'' method consists of first inferring a ``priming variable'', typically a coarse estimate of the label, from the key inputs alone. Then, the target DNN which is supplied with this priming variable alongside the full input during training becomes much more likely to avoid shortcuts and ``learn the right thing.'' 

PrimeNet is very simple to implement, can be trained end-to-end, and yields large gains across benchmark tasks for several practical applications ranging from out-of-domain image classification on NICO~\citep{he2021towards} to imitation learning for controlling robots in MuJoCo~\citep{todorov2012mujoco} and autonomous cars in CARLA~\citep{Dosovitskiy17}. Finally, we argue from recently developed theories of neural network learning that shortcuts are caused by optimization biases that prefer \textit{simpler} solutions even if they may be trivial or wrong, and that our priming approach works by creating an even simpler yet approximately correct solution within the optimization landscape.

%% file: 3_method.tex
\begin{figure*}[t]
    \centering
    \includegraphics[height=0.2\linewidth]{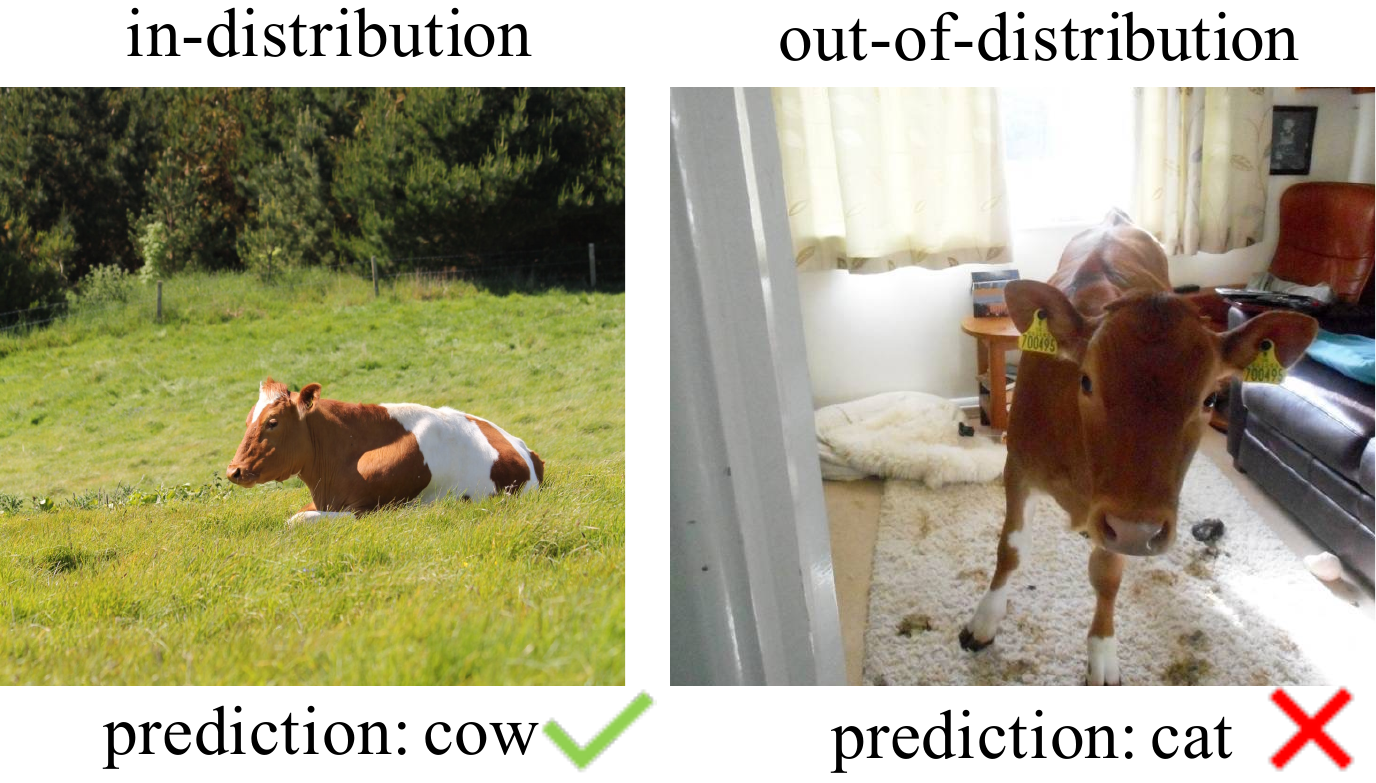}
    \quad
    \quad
    \includegraphics[height=0.2\linewidth]{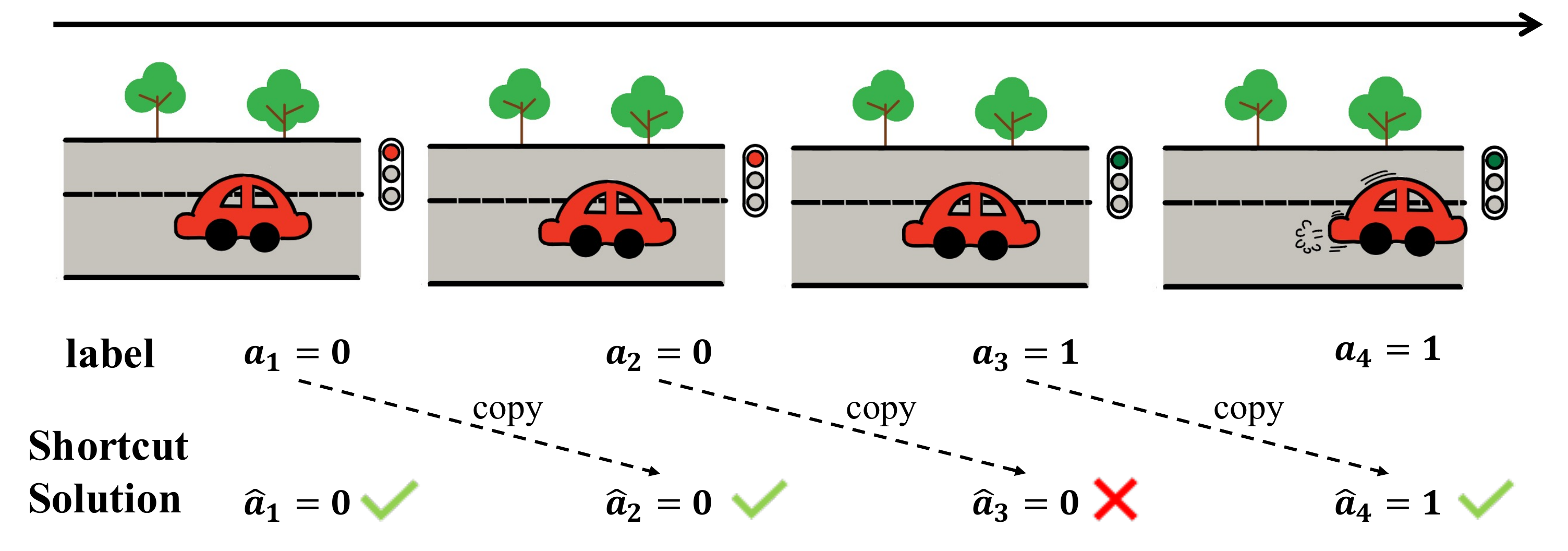}

    \caption{ \textbf{(Left)} An example of shortcut learning in image classification in NICO~\citep{he2021towards}. The trained model successfully predicts the ``cow" class in the image where the background is grass, while it classifies the cow at home poorly. It indicates that the model recognizes the cow according to the backgrounds rather than the objects. \textbf{(Right)} An example of copycat shortcuts in autonomous driving. The copycat policy learns to cheat by copying from the previous action during training. Therefore, during online testing, when the traffic light turns green, the copycat policy ignores the traffic light and copies the previous action, i.e. staying stationary.}
    \label{fig:cow-example}\label{fig:copycat}
\end{figure*}

\section{Preliminaries: Shortcuts}\label{sec:preliminaries}

As introduced above, many surprising error tendencies in deep neural networks arise from strategies that are ``superficially successful'' (under training circumstances), but fail catastrophically under slightly different circumstances. Following \citet{geirhos2020shortcut}, we use the term ``shortcut issue'' to refer to such errors. 

Formally, let $\mathcal{P}_{in}(\boldsymbol{x},\boldsymbol{y})$ denote a joint probability distribution over $\mathcal{X}$ and $\mathcal{Y}$ from which training data is drawn I.I.D. i.e. $\mathcal{D}_{in}=\{(\boldsymbol{x_i},\boldsymbol{y_i})\}_{i=1}^{n} \sim \mathcal{P}_{in}$. 
Let $\mathcal{P}_{out}(\boldsymbol{x},\boldsymbol{y})$ denote a different distribution from which the out-of-distribution (O.O.D.) testing set is similarly drawn, i.e. $\mathcal{D}_{out}=\{(\boldsymbol{x_i},\boldsymbol{y_i})\}_{i=1}^{m} \sim \mathcal{P}_{out}$. 
A neural network $f_{\theta}(\cdot)$, parameterized by $\theta$, is trained by SGD with a loss function $l(\cdot, \cdot)$ on the training set $\mathcal{D}_{in}$.  
Let $\theta^{*}$ denote the intended solution, which can successfully generalize to the O.O.D. distribution, i.e. $\theta^{*} = \arg\min_{\theta}\mathbb{E}_{\mathcal{P}_{in}+\mathcal{P}_{out}}[l(f_{\theta}(\boldsymbol{x}),\boldsymbol{y})]$. 
For simplicity, we use $L_{\mathcal{P}}(\theta)$ to denote the population loss of solution $\theta$ on distribution $\mathcal{P}$.
Shortcut learning, or the shortcut issue, refers to learning solutions $\theta_s$ that perform well under the training distribution, i.e. $L_{\mathcal{P}_{in}}(\theta_{s}) \approx L_{\mathcal{P}_{in}}(\theta^{*})$, but generalizes very poorly to O.O.D. data, i.e.  $L_{\mathcal{P}_{out}}(\theta_{s}) \gg L_{\mathcal{P}_{out}}(\theta^{*})$.

Shortcut issues have been observed in many applications, including computer vision~\citep{beery2018recognition,GeirhosRMBWB19}, NLP~\citep{niven2019probing,mccoy2019right}, imitation learning~\citep{bansal2018chauffeurnet,codevilla2019exploring,chuan2020fighting} and reinforcement learning~\citep{amodei2016concrete,zhang2021learning}. 
We highlight image classification and imitation learning shortcuts below, which we will later validate our proposed approach on.

\textbf{Image classification shortcuts:} 
Consider a DNN-based animal image classifier. Cows in training typically have grass backgrounds, and the classifier may easily learn to rely on grass as an important ``shortcut'' cue for the ``cow'' label. Indeed, such a solution would even generalize to in-distribution test data, but fail on in-the-wild distributionally shifted images with cows at home (see Figure~\ref{fig:cow-example}).   
Many such shortcuts have recently drawn attention in the image recognition literature~\citep{beery2018recognition,rosenfeld2018elephant,buolamwini2018gender}.

\textbf{Imitation learning shortcuts:} Next, consider the task of learning to control a robot or autonomous car from observing expert demonstrations of actions $\boldsymbol{a}_t$ corresponding to sensory observations (such as camera image streams) $\boldsymbol{x}_t$ at the same time $t$. ``Behavioral cloning'' (BC) treats this as supervised learning to regress $\boldsymbol{a}_t$ from recent observation histories $[\boldsymbol{x}_{t-k}, \cdots, \boldsymbol{x}_t]$. In theory, the inclusion of older observations as inputs allows the imitator to  compensate for partial observability: for example, a pedestrian occluded in the current camera image from a car dashboard might have been visible earlier. However, the core problem of BC lies in generalizing beyond expert (hence in-distribution) data. Indeed, many previous works~\citep{chuan2020fighting, muller2006off,de2019causal,bansal2018chauffeurnet,codevilla2019exploring,chuan2021keframe,ortega2021shaking} find that BC from observation histories, much like the image classifier above, yields better training and validation losses on expert data, but fails catastrophically when executed on a robot or car. 
Like the classifier mistaking the grass for the cow, DNN imitators in these situations frequently learn shortcuts that simply copy the last performed action rather than respond to new road images or other observations. This happens because experts usually act smoothly, and adjacent actions in training data are nearly always identical. This ``copycat'' shortcut can cause curious phenomena like the ``inertia problem'' (see Figure~\ref{fig:copycat}) where cars that once come to rest never move again!

\section{Method: Priming DNNs To Avoid Shortcuts}

We now describe the core contribution of this paper: an easy-to-implement solution to avoid DNN shortcuts in settings where auxiliary domain knowledge about ``key inputs'' is available. Section~\ref{sec:motivation} sets up our key idea, \ref{sec:primenet} describes the algorithm in detail, \ref{sec:theory} provides theoretical support, and \ref{sec:method:two_applications} discusses two example applications.

\subsection{Motivation}\label{sec:motivation}

Out-of-distribution generalization for supervised learning is hard, yet humans often manage to generalize far beyond the training data~\citep{geirhos2018generalisation} by exploiting additional knowledge beyond the labels. 
Recall the example from the introduction of the parent teaching the child to recognize zebras. Besides ``labeling'' the scene as containing a ``zebra'', the parent also points to the animal and its stripes. Without this auxiliary information, it would be a much harder task for the child to acquire the correct concept of a zebra. However, in the common supervised machine learning setup, no such additional information is available; instead, the supervision consists only of labeled examples.
We propose to expand this supervision by providing auxiliary knowledge about pertinent ``key inputs'' to DNNs, to help them to avoid shortcuts and learn the right thing. 

To use such auxiliary information well, we draw inspiration from cognitive science. Cognitive scientists have observed a phenomenon called ``priming'': exposing humans to one stimulus influences their response to a subsequent stimulus, without conscious guidance or intention~\citep{weingarten2016primed,bargh2014mind}. For example, people who were recently exposed to words associated with the elderly (e.g., retirement) begin to walk more slowly than before~\cite{bargh1996automaticity}.
We propose to prime DNNs during training with the known key inputs to coax them to discover solutions that rely more on that information, matching our knowledge about the correct solution. Next, we will describe this approach, \textbf{PrimeNet}, in more detail.

\subsection{PrimeNet}\label{sec:primenet}
As motivated above, PrimeNet exploits domain knowledge of portions of the full input $x$ that are most likely to contain task-relevant information. While specifying such knowledge for each example would be cumbersome and impractical, we observe that it is often easy to specify this knowledge at a task level. For example, for image classification tasks, the foreground objects, identified perhaps by a generic object detector or salient foreground segmenter, could be specified to the learner as being the most important. In a partially observed sequential control task like vision-based driving, one might specify the most recent frame as likely to contain the most pertinent information. We call these the ``key inputs'', denoted as
$k(x)$, where the $k(\cdot)$ is a specified process or function that extracts them from the full inputs $x$. The full task specification for PrimeNet thus augments the standard training dataset with this function $k(.)$.

\begin{figure}[htb]
    \centering
    \includegraphics[width=\linewidth]{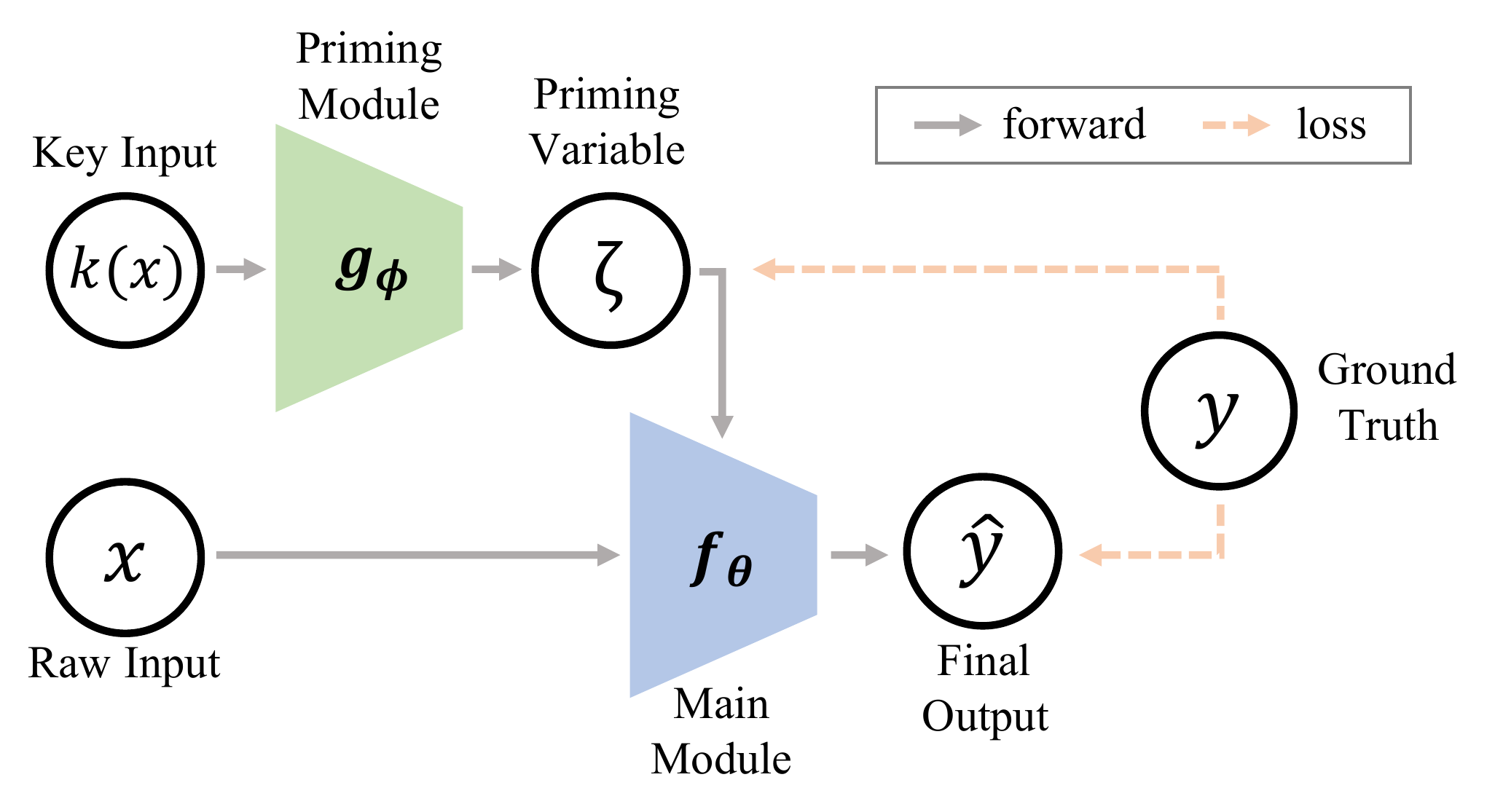}
    \caption{The PrimeNet framework. From the full input $x$, the key input $k(x)$ is extracted using domain knowledge specified through a function $k(.)$. This is then used to ``prime'' the main module $f_\theta$ towards good solutions during training. 
    }
    \label{fig:prime_arch}
\end{figure}

As shown in Figure~\ref{fig:prime_arch}, the PrimeNet framework contains two modules: a \textbf{priming module} $g_{\phi}$ and the \textbf{main module} $f_{\theta}$. 
In the forward pass, the priming module $g_{\phi}$ first computes a coarse estimate $\zeta$ of the target label based on the key inputs $k(x)$, i.e. $\zeta = g_{\phi}(k(x))$. We call this key-input-based coarse estimate $\zeta$ the \textbf{priming variable}. Then, the main module $f_{\theta}$ receives $\zeta$ as input alongside the full input $x$, and produces the final output $\hat{y}=f_{\theta}(x,\zeta)$. 

We jointly train the priming module $g_{\phi}$ and main module $f_{\theta}$ end-to-end on the in-distribution training set $\mathcal{D}_{in}={(x_{i},y_{i})}_{i=1}^{n}$ by minimizing standard loss functions $l(\cdot, \cdot)$ as appropriate to the task (e.g., the cross-entropy loss for classification and MSE loss for regression) computed on the outputs of both the priming module (i.e., the priming variable, which we train to be a coarse estimate of the output) and the main module (i.e., the final output).  
\begin{equation*}
\begin{aligned}
    \phi^{*} &= \arg\min_{\phi}\frac{1}{n}\sum_{i=1}^{n}l(g_{\phi}(k(x_{i})),y_{i}) \\
    \theta^{*} &= \arg\min_{\theta}\frac{1}{n}\sum_{i=1}^{n}l(f_{\theta}(x_{i},\zeta_{i}),y_{i})
\end{aligned}
\end{equation*}
In our implementation, we concatenate $\zeta$ with intermediate activations in the DNN $f_{\theta}$ and feed them into the following layers together. In this way, the main module is still free to learn a solution that relies on information in the full input that is not available in the key input. This is a necessary property: recall that key inputs merely provide good \textit{summaries} of task-relevant information, and may not be comprehensive. For example, looking at the background often provides helpful contextual cues when classifying images~\cite{oliva2007role}, and remembering older observations may be necessary to account for an occluded pedestrian during driving~\cite{bansal2018chauffeurnet}. 

This brings us to the question: given that priming does not restrict the hypothesis space and the DNN \textit{can} still represent the same shortcut solutions as before, how does priming improve DNN training at all, as our results will suggest in Sec~\ref{sec:experiments}?
We argue that since $\zeta$ itself is a coarse solution for the task, it introduces a simple and desirable shortcut towards low training losses which sets the optimization process for $f_\theta$ ``on the right track'', away from any catastrophically wrong shortcut solutions.
Section~\ref{sec:theory} formalizes this intuition.

\subsection{Theoretical Justification}\label{sec:theory}

To explain why PrimeNet takes the shortcut from the priming variable $\zeta$ rather than the undesired shortcuts, we derive a property of the neural networks that different inputs guide the optimization to different solutions.

\input{3_propostition}

\textbf{Remark.}
What does Proposition~\ref{prop: early} say about why PrimeNet works? If the input of the main module $f_{\theta}$ is $[x,\zeta]$, then the DNN will prefer simple functions (such as linear functions) of
$x$ and $\zeta$ even in the O.O.D. region, and changes in $\zeta$ may significantly change this preference, even though $\zeta$ is computed from the full input $x$ and therefore cannot introduce any new information. Thus, setting $\zeta$ to represent coarse output estimates based on key input domain knowledge encourages the DNN to find solutions that also rely on the key input. 

\subsection{Applying PrimeNet}\label{sec:method:two_applications}
Recall the two shortcut issue examples from Section \ref{sec:preliminaries}. For the image classification problem, shortcuts are caused by background distractions, and we consequently set the key input function $k(.)$ to be an image patch crop from unsupervised saliency detection~\citep{qin2019basnet}. This provides a background-free image patch for priming without the risk of introducing shortcuts into the extraction of the priming variable $\zeta$ itself. In the imitation setting, shortcuts are caused by the implicit previous action in the historical observations. To remove this shortcut, we propose to use the most recent frame as the priming input, i.e. $k(x)=x_t$. Once again, this might lose some information, but it can serve as the basis for a coarse action estimate that is free from shortcuts. See Section~\ref{sec:experiments} and Appendix~\ref{app:nico-exp-detail} and Appendix~\ref{app:carla-exp-detail} for more details.

%% file: 3_propostition.tex
\begin{proposition}
\label{prop: early}
Consider two functions $h$ and $s$ within the hypothesis space of a neural net, where $h$ is linear with respect to the input.
Further, suppose that:
\vspace{-0.1in}
\begin{itemize}[leftmargin=1cm]
  \setlength\itemsep{0.1em}
    \vspace{-0.05in}
    \item[\textbf{[A1]}] In the training region, functions $h$ and $s$ are both close to the ground truth.
    \vspace{-0.1in}
    \item[\textbf{[A2]}] In the out-of-distribution testing region, functions $h$ and $s$ are far apart.
\end{itemize}
\vspace{-0.1in}
Then there exists a training iterate, such that for a neural tangent kernel $f_{NTK}$ trained at such iterate, the following statements (C1-C2) hold with high probability under some mild conditions:
\vspace{-0.1in}
\begin{itemize}[leftmargin=1cm]
\setlength\itemsep{0.1em}
\vspace{-0.05in}
\item[\textbf{[C1]}] In the in-distribution region, $f_{NTK}$ reaches small training error.
\vspace{-0.1in}
\item[\textbf{[C2]}]  In the out-of-distribution region, $f_{NTK}$ approximates $h$ well but stays far from $s$. 
\end{itemize}
\vspace{-0.1in}
\end{proposition}

We defer the formal statement and proof to Appendix~\ref{app:theoremFormal}. Broadly, this proposition states that neural network training, approximated by neural tangent kernels~\citep{jacot2018neural}, is biased towards solutions that are linear functions of the inputs.
In the training region, both the function $h$ which is linear in the input features, as well as a non-linear function $s$ approximate ground truth $\y$ well (Assumption~A1). 
Therefore, DNN training may recover either $h$ or $s$ as valid solutions.
However, 
\ref{prop: early} demonstrates that in the O.O.D. region, neural networks indeed prefer the simple function $h$, which is linear in the input features.
The proof starts from \citet{DBLP:conf/nips/HuXAP20}, which shows that neural networks during early training approximate a linear model, and we further extend the conclusion to the O.O.D.~region, providing theoretical justification for using priming to solve shortcut problem (see below).

%% file: 4_experiments.tex
\section{Experiments}
\label{sec:experiments}

We conduct experiments on three sets of tasks: a toy regression experiment, an image classification task and two behavioral cloning tasks on autonomous driving and robotic control to verify our arguments and the proposed method. The toy experiment is to validate the Proposition~\ref{prop: early}. The realistic image classification and behavioral cloning tasks are designed to verify that our method can resolve shortcut issues and evaluate it against previous state-of-art methods. Finally, we conduct ablation and analysis studies on these two tasks to study our method more closely. 

\begin{table}[ht]
\caption{RMSEs for the 1-D regression experiment.}
\label{tab:toy-experiment}
\begin{center}
\begin{scriptsize}
\begin{sc}
\begin{tabular}{c|cccc}
\toprule
& \multicolumn{2}{c|}{I.I.D.} &
\multicolumn{2}{c}{O.O.D.} \\
\cmidrule(lr){2-3} \cmidrule(lr){4-5}
$\zeta$ value &  $f_{1}$  &  \multicolumn{1}{c|}{$f_{2}$}   &  $f_{1}$  &  $f_{2}$ \\
\midrule
0    &  0.101  &  0.121  &  18.126  &  8.838  \\
$x^{4}$    &  0.100  &  0.122  &  9.062  &  \textbf{0.258}  \\
$x^{5}$    &  0.100  &  0.122  &  \textbf{0.243}  &  9.075  \\
\bottomrule
\end{tabular}
\end{sc}
\end{scriptsize}
\end{center}
\end{table}

\subsection{One-Dimensional Regression}
We design a toy 1-D regression experiment on synthetic data to empirically validate Proposition~\ref{prop: early}.
To construct a scenario with multiple local optimal solutions on the training set and with the challenge of O.O.D. generalization, we design two functions: $f_{1}(x)=1.5x^5+2x+\epsilon$ and $f_{2}(x)=1.5x^4+2x+\epsilon$, where $\epsilon\sim N(0,0.1)$ is an additive Gaussian noise. These two functions are close to each other in the training region $[0,1]$, and significantly different in the testing region $[1,2]$. We uniformly sample 1000 training pairs $(x_{i}, f_{1}(x_{i}))$ in the training region $x_{i} \in U(0,1)$. We train a two-layer MLP to fit the training data with the input $[x,\zeta]$, where $[\cdot,\cdot]$ means concatenation operation and $\zeta$ is the priming variable. We train three models with $\zeta$ equal to 0, $x^{4}$ or $x^{5}$ respectively. We test each trained model on two testing regions, i.e. the in-distribution region $[0,1]$ and O.O.D. region $[1,2]$, and with two reference functions as priming variables $\zeta$, i.e. $f_{1}$ and $f_{2}$. We want to study which solution the MLP will converge to, when given different $\zeta$. The accuracy on the O.O.D. region will indicate what function the neural network has learned during training.

Table~\ref{tab:toy-experiment} reports the RMSE value of the trained models against $f_1$ and $f_2$ on in-distribution and out-of-distribution testing sets. All models fit the training data perfectly regardless of the choice of priming variable $\zeta$, and the RMSEs on $f_{1}$ and $f_{2}$ are close in the in-distribution region $[0,1]$ because the two functions are very similar in this range. 
When testing the models in the O.O.D. region $[1,2]$, the model fits $f_{2}$ better when primed on $x^4$, and fits $f_1$ better if primed on $x^{5}$. This is true even though the labels in the training data were generated by $f_1$. 
This shows that the priming variable $\zeta$ can significantly influence DNN training towards simpler functions of $[\zeta, x]$, consistent with Proposition~\ref{prop: early}. 
We visualize the learned functions in Appendix~\ref{app:toy-exp-curve}.

\subsection{Image Classification}

O.O.D. image classification is a challenging task that has attracted attention in computer vision in recent years~\citep{arjovsky2019invariant,krueger2021out,xu2021fourier,wang2021causal}. We evaluate PrimeNet on an O.O.D. image classification benchmark.

NICO~\citep{he2021towards} is an image classification dataset designed for O.O.D. settings. In total, it contains 19 classes, 188 contexts and nearly 25,000 images. Besides the object label (cat, dog, cow, etc.), each of the images is also labeled with a context label (at home, on beach, on grass, etc.). Thus, it is convenient to design the distribution of data by adjusting the proportions of specific contexts for training and testing images. We use the animal subset of NICO with 10 categories, and each class has 10 contexts. 
Following \citet{wang2021causal}, we construct a challenging O.O.D. setting: 1) the training dataset has only 7 contexts for each object category and the testing dataset has all the 10 contexts to evaluate the \textbf{zero-shot} generalization capability of the models (see the examples in Figure~\ref{fig:cow-example} (left)); 2) the context labels in the training set are in \textbf{long-tailed} distribution, which makes it more difficult to train an unbiased classifier.
We employ ResNet-18~\citep{he2016deep} as the backbone network for all the methods. For our method, we use a weight shared ResNet-18 as $g_{\phi}$ and $f_{\theta},$ and utilize the unsupervised saliency detection model BASNet~\citep{qin2019basnet} to crop the salient areas of the full images as $k(x)$. See Appendix~\ref{app:nico-exp-detail} for the architecture, implementation and training details.

\subsubsection{Baselines}
For image classification, we extensively compare PrimeNet with four classes of methods: a set of vanilla baselines designed by us, and three sets of methods to improve generalization using debias, image augmentation and intervention techniques.

\textbf{Vanilla baselines.}
We design three vanilla baselines with the ResNet18 model: 1) \textbf{Vanilla ResNet18}~\citep{he2016deep}: directly adopting the ResNet18 model to classify the images; 2) \textbf{Key-Input-Only} baseline: using the key input $k(x)$ as the input of ResNet18; 3) \textbf{Average-Ensemble} baseline: training ResNet18 on an expanded dataset containing the original images $x$ as well as the extracted key input regions $k(x)$, and then averaging the output logits for $x$ and $k(x)$ for classification during testing.

\textbf{Debias methods.} 
We compare our method with two SOTA debias methods: \textbf{RUBi}~\citep{cadene2019rubi}, \textbf{ReBias}~\citep{bahng2020learning} and \textbf{StableNet}~\citep{zhang2021deep}. RUBi learns a biased model with the biased dataset and then trains an unbiased model by re-weighting according to the predicted logits of the biased model. ReBias firstly train a small biased model with the biased dataset and regularizes the main model to be statistically independent from it. StableNet removes spurious correlations by reweighting training samples to get rid of feature dependencies.

\textbf{Data augmentation methods.}
We adopt two commonly used data augmentation methods, \textbf{Mixup}~\citep{zhang2017mixup} and \textbf{Cutout}~\citep{devries2017improved}. Mixup linearly interpolates the inputs and labels of two random training samples, thus extending the training distribution. Cutout randomly masks out a square patch on the full image to avoid overfitting to the contexts.

\textbf{Causal inference methods.}
We also compare to two SOTA causal methods: \textbf{IRM}~\citep{arjovsky2019invariant} and \textbf{Caam}~\citep{wang2021causal}. IRM proposes to learn an invariant representation which gets optimal classification performance across different environments (contexts) to remove the effect of the spurious shortcut correlations. It requires the environment labels. Caam improves IRM by automatically getting the environment partitions in the training set and utilizes IRM to boost the performance.

\subsubsection{Results}

Table~\ref{tab:nico-results} shows the classification accuracy on test sets of all the methods. PrimeNet achieves significantly better performance than the baselines across all the settings. 
PrimeNet (ours) and CAAM significantly improve the in-domain accuracy (as well as OOD), and ours is better than CAAM on both. We believe this happens because NICO has relatively small training datasets with incomplete coverage of the training distribution, so even in-domain generalization can sometimes benefit from removing shortcuts. Other methods (e.g. Average-Ensemble and StableNet) improve the OOD generalization while harming the in-domain performance.

\begin{table}[t]
\caption{Image classification accuracies on the NICO dataset. Baseline scores from \citet{wang2021causal}. ``-" means the value is neither reported in \citet{wang2021causal} nor reproduced by us because we do not have the source codes. \footnotemark}
\label{tab:nico-results}
\begin{center}
\begin{scriptsize}
\begin{sc}
\begin{tabular}{c|cc}
\toprule
Method &  in-domain test  &  ood test      \\
\midrule
vanilla ResNet18    &  66.11  &  42.61  \\
key-input-only    &  62.78  &  47.54  \\
average-ensemble    &  63.33  &  47.69  \\
Rubi~\citep{cadene2019rubi}    &  -  &  44.37  \\
ReBias~\citep{bahng2020learning}    &  -  &  45.23  \\
CutOut~\citep{devries2017improved}    &  -  &  43.77  \\
MixUp~\citep{zhang2017mixup}    &  62.78  &  41.46  \\
IRM~\citep{arjovsky2019invariant}    &  -  &  41.46  \\
StableNet~\citep{zhang2021deep}    &  63.33  &  43.62 \\
Caam~\citep{wang2021causal}    &  70.00  &  46.62  \\
\midrule
PrimeNet (ours)    &  71.11  &  \textbf{49.00}  \\
\bottomrule
\end{tabular}
\end{sc}
\end{scriptsize}
\end{center}
\end{table}

\footnotetext{Note that the test accuracy of StableNet~\citep{zhang2021deep} is lower than that in the original paper because we follow \citet{wang2021causal} to construct a more difficult dataset based on NICO with long-tailed and zeroshot properties.}

As for OOD performance, compared to vanilla ResNet18, PrimeNet improves accuracy by more than 6\%, illustrating that our method successfully alleviates the shortcuts from the background contexts, and thus gets better generalization performance. We note that although the Key-Input-Only method already performs better than previous methods, our PrimeNet further improves on top of it, showing that PrimeNet manages to effectively integrate information from non-key areas, unlike Key-Input-Only. Compared with the SOTAs of debias (Rubi, ReBias and StableNet), data augmentation (Cutout and Mixup) and intervention (IRM and Caam) methods, our method introduces an alternative inductive bias, i.e. key input from unsupervised saliency detection. This does not require any additional supervision, relying only on the domain knowledge that foreground objects contain the most pertinent information for this task. This validates that saliency-based key input priming is an effective and practical inductive bias for resolving image classification shortcuts.

\subsection{Behavioral Cloning For Imitation}

The O.O.D. generalization problem is the main bottleneck of behavioral cloning, which is widely recognized over many decades~\citep{muller2006off,Ross2011,bansal2018chauffeurnet,chuan2020fighting,spencer2021feedback}. We evaluate PrimeNet on two imitation learning tasks.

\textbf{CARLA.}
CARLA is a photorealistic urban autonomous driving simulator~\citep{Dosovitskiy17}, and it is a commonly adopted testbed for imitation learning~\citep{codevilla2018end,codevilla2019exploring,chen2020learning,chuan2021keframe,prakash2021multi}. We use the CARLA100 dataset to train all methods and evaluate them in the \textit{Nocrash} benchmark~\citep{codevilla2019exploring}. 
Following~\citet{chuan2021keframe}, the POMDP observations don't include the vehicle speed, and imitation learners can instead use past video frames to prescribe driving actions.
We train all the methods three times from different random initializations to account for variance~\citep{codevilla2019exploring}. We use CILRS~\citep{codevilla2019exploring} as the backbone, set the length of input observation history to $7$, and stack the sequential frames along the channel dimension as the model input as in \citet{bansal2018chauffeurnet,chuan2021keframe}. 
We report the mean and standard deviation of two metrics: \%success and \#timeout. \%success is the number of episodes that are fully completed out of 100 pre-designed evaluation routes. \#timeout counts the times that the agent fails to reach the destination despite no collision within the specified time. Timeout is usually caused by unsuccessful starts, wrong routes or traffic jams. We report more evaluation metrics, and implementation details in Appendices~\ref{app:more-carla-results} and \ref{app:carla-exp-detail}.

\textbf{MuJoCo.} 
We evaluate our method in three standard OpenAI Gym MuJoCo continuous control environments: Hopper, Ant and HalfCheetah. We generate expert data from a PPO~\citep{schulman2017proximal} policy (10k samples for Ant and Walker2D, and 20k for Hopper). To simulate partial observations, we add Gaussian noise ${N}(0, \sigma^{2})$ with $\sigma=0.2$ to joint velocities. We stack 2 frames to form the observation history, and use 2-layer MLPs as policy network backbones. We train all methods with 3 random initializations and report mean and standard deviation of rewards. See Appendix~\ref{app:mujoco-exp-detail} for network architecture and training details.

\subsubsection{Baselines}
For BC, we compare against three groups of methods:

\textbf{Vanilla baselines.} We retain the vanilla baselines from image classification, training them for BC. Note that \textbf{vanilla-BC} and \textbf{key-input-only} corresponds to behavioral cloning from observation histories and single observation (BC-OH and BC-SO respectively) as studied in prior works~\cite{chuan2020fighting,chuan2021keframe}. We use CILRS~\citep{codevilla2019exploring} and two-layer MLPs as the policy network backbone in place of ResNet18 for CARLA and MuJoCo respectively.

\textbf{Previous methods tackling BC shortcuts.}
We compare PrimeNet with three previous solutions to the shortcuts in BC: \textbf{fighting copycat agents (FCA)}~\citep{chuan2020fighting}, \textbf{KeyFrame}~\citep{chuan2021keframe} and \textbf{History-Dropout}~\citep{bansal2018chauffeurnet}. FCA proposes to resolve the shortcut issue with adversarial training, which removes information about the previous action $a_{t-1}$. KeyFrame tackles the shortcut in BC from an optimization perspective: it up-weights the datapoints at action change-points, which is defined as an action prediction module's failure timesteps. History-Dropout introduced a dropout on the observation history to randomly erase the channels of historical frames. 

\textbf{Online ``upper bound'' solution.} 
We also compare with an online imitation learning method, \textbf{DAGGER}~\citep{Ross2011}. DAGGER is a widely used method to mitigate domain shift issues in imitation learning, but requires online environmental interaction with a queryable expert. We note that our method does not require the online expert, and thus DAGGER is an ``upper bound'', rather than a baseline. In our experiments, DAGGER uses 150k supervised interaction steps for CARLA and 8k for MuJoCo.

\subsubsection{Results}

\begin{table*}[ht]
\caption{Behavioral cloning results on CARLA and MuJoCo.}
\label{tab:bc-results}
\begin{center}
\begin{scriptsize}
\begin{sc}
\begin{tabular}{c|cc|cccc}
\toprule
& \multicolumn{2}{c|}{carla results} &
\multicolumn{3}{c}{mujoco rewards} \\
\cmidrule(lr){2-3} \cmidrule(lr){4-6}
method &  \%success  &  \multicolumn{1}{c|}{\#timeout}   &  hopper  &  ant  &  halfcheetah  \\
\midrule
vanilla bc    &  34.1 $\pm$ 7.5  &  36.1 $\pm$ 14.5  &  628 $\pm$ 99 & 2922 $\pm$ 1266 &  639 $\pm$ 121 \\
key-input-only    &  13.1 $\pm$ 1.8  &  \textbf{11.1 $\pm$ 2.9}  & 589 $\pm$ 94 & 4198 $\pm$ 433  & 489 $\pm$ 77  \\
average-ensemble  &  41.7 $\pm$ 3.1  &  15.0 $\pm$ 0.8  & 504 $\pm$ 47  & 4659 $\pm$ 396 &729 $\pm$ 50 \\
\textbf{PrimeNet (ours)}     &  \textbf{49.3 $\pm$ 3.6}  &  12.0 $\pm$ 1.9  & \textbf{1124 $\pm$ 135} & \textbf{4798 $\pm$ 304} &  \textbf{1448 $\pm$ 74} \\
\midrule
fca~\citep{chuan2020fighting}      &  31.2 $\pm$ 5.2  &  35.3 $\pm$ 9.6  & 831 $\pm$ 108 & 3727 $\pm$ 926 &   1148 $\pm$ 81 \\
Keyframe~\citep{chuan2021keframe}  &  41.9 $\pm$ 6.2  &  24.8 $\pm$ 7.9  & 696 $\pm$ 28 & 2930 $\pm$ 1321 & 1062 $\pm$ 127 \\
history-dropout~\citep{bansal2018chauffeurnet}  &  35.6 $\pm$ 3.5  &  20.3 $\pm$ 5.6  & 539 $\pm$ 33  &  4069 $\pm$ 517& 1215 $\pm$ 70 \\
\midrule
dagger~\citep{Ross2011}  &  42.7 $\pm$ 5.7  &  23.0 $\pm$ 7.1  & 2383 $\pm$ 294 & 4097$\pm$ 418 &1842 $\pm$ 10 \\
\bottomrule
\end{tabular}
\end{sc}
\end{scriptsize}
\end{center}
\end{table*}

\textbf{CARLA.}
Table~\ref{tab:bc-results} shows results on the hardest benchmark with the densest traffic, \textit{Nocrash-Dense}. Key-input-only performs poorly, and is even worse than vanilla BC on \%success. It sees only the last frame and has no way to judge its own speed, so it most commonly fails by accelerating at all times on straight roads, resulting in speeding and collisions.
Vanilla BC has access to past frames, but suffers from very high \#timeout rates due to starting problems: once the car stops, say, at a traffic light, vanilla BC often remains stationary until timeout. This high \#timeout is known to be a typical copycat shortcut symptom~\citep{chuan2020fighting,chuan2021keframe,bansal2018chauffeurnet} for BC from observation histories.

It is thus especially interesting that PrimeNet reduces the \#timeout to $12.0$, very close to key-input-only. PrimeNet also outperforms all baselines on \%success, remarkably even including the ``upper bound'' DAGGER.
Appendix~\ref{app:more-carla-results} shows results on \textit{Nocrash-Empty} and \textit{Nocrash-Regular}.

\textbf{MuJoCo.}
Table~\ref{tab:bc-results} also shows results on Hopper, Ant, and HalfCheetah. 
Here, vanilla BC only performs slightly better than key-input-only on Hopper and HalfCheetah, and does worse than key-input-only on Ant, due to the copycat shortcut. PrimeNet beats out all baselines, including prior approaches addressing BC shortcuts (FCA, KeyFrame and History-Dropout), in all 3 environments. The upper bound DAGGER approach does significantly better than all methods on Hopper and Half-Cheetah, but Key-Input-Only and PrimeNet beat it on Ant.

\subsection{Ablation Study} \label{sec:ablation}

\begin{table}[ht]
\caption{The ablation study results on NICO and CARLA.}
\label{tab:ablation}
\begin{center}
\resizebox{.5\textwidth}{!}{
\begin{scriptsize}
\begin{sc}
\begin{tabular}{c|c|cc}
\toprule
& \multicolumn{1}{c|}{NICO} &
\multicolumn{2}{c}{CARLA} \\
\cmidrule(lr){2-2} \cmidrule(lr){3-4}
method &  \multicolumn{1}{c|}{test acc.}   &  \%success  &  \#timeout \\
\midrule
\textbf{PrimeNet (ours)}    &  \textbf{49.00}  &  \textbf{49.3}   &  \textbf{12.0}  \\
\midrule
ResNet-early-feature as $\zeta$    &  45.00  &  39.8  & 28.1  \\
ResNet-late-feature as $\zeta$    &  48.62  &  44.8  &  15.9 \\
\midrule
No-Key-Input    &  43.92  &  37.1  &  28.9 \\
$k(x)$-for-Both-Modules &  47.92  &  13.8  &  11.9 \\
\bottomrule
\end{tabular}
\end{sc}
\end{scriptsize}
}
\end{center}
\vskip -0.2in
\end{table}

We now study PrimeNet in more detail through ablations.

\textbf{Priming variable selection.} Should the priming variable be the output of priming module $g_\phi$, as we have been using, or should it rather be an intermediate representation of the key input from $g_\phi$? We evaluate an early/shallow (layer2) and a late/deep feature (last layer) from $g_\phi$ (in place of the output) as the priming variable to be fed to the main module.
Table~\ref{tab:ablation} shows a clear trend: shallower priming variables are always worse, and the output from $g_\phi$ performs best. This suggests that shallower features do not as effectively create good priming shortcuts to prevent bad shortcut learning.

\textbf{Role of key inputs. }
In Table~\ref{tab:ablation}, we report the results of using our architecture with no key input (with $x$ as the input to both trunks), and an architecture with key input $k(x)$ for both modules.
Under both conditions, the performance on both NICO and CARLA becomes worse,  with no-key-input even close to the vanilla baseline. This indicates that the specific design choices of PrimeNet are important to help avoid shortcuts.
Appendix~\ref{app:more-ablation} shows more ablations.

\subsection{Analysis} \label{sec:analysis}
\begin{figure*}[t]
    \centering
    \includegraphics[width=0.7\textwidth]{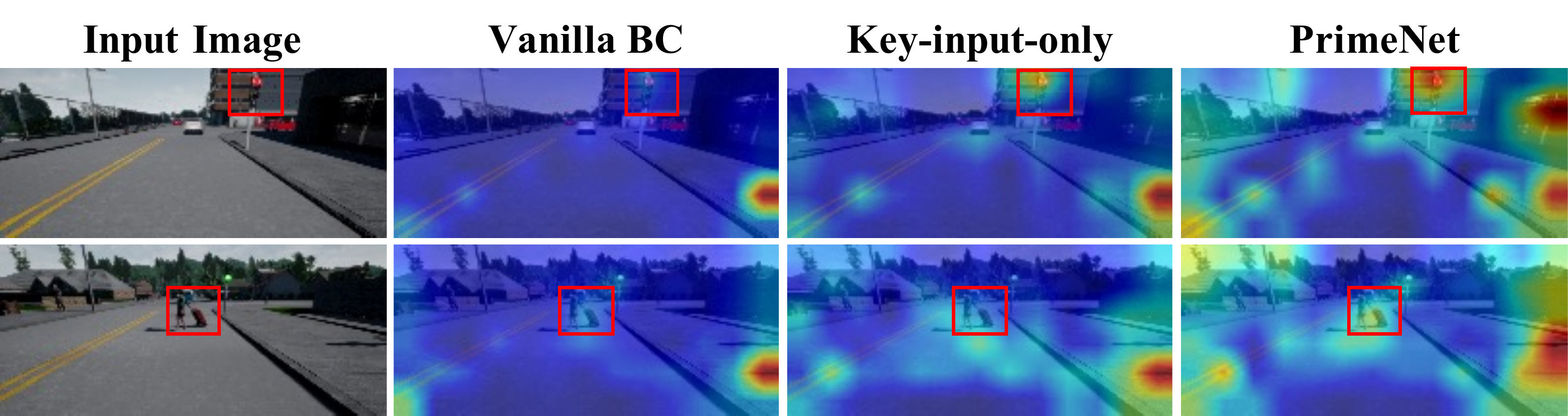}
    \caption{The activation maps are extracted from the Layer3 in the ResNet34 of the perception module of the backbone CILRS~\citep{codevilla2019exploring}. There are two scenarios of slowing down in front of a red traffic light (top) and a pedestrian (bottom). The red areas in the red boxes are where the policies should pay attention to. We can see that the key-input-only model and our PrimeNet correctly focus on the traffic light and the pedestrian while the vanilla BC pays much less attention to the pedestrian and even ignores the traffic light.
    }
    \label{fig:heatmap}
\end{figure*}

We use the activation map~\citep{muhammad2020eigen} of ResNet Layer3 on CARLA to illustrate the visual cues that are attended to by different models. 
As shown in Figure~\ref{fig:heatmap}, key-input-only model and our PrimeNet correctly notice the red traffic light and the pedestrian. In contrast, vanilla BC shows symptoms of the shortcut problem, paying little attention to the pedestrian and ignoring the traffic light, eventually causing traffic light violation and collision. PrimeNet thus learns to attend to the correct cues in the observations, rather than rely on shortcuts. The activation maps of NICO experiment can be found in Appendx~\ref{app:nico-visualization}. Appendix~\ref{app:more-analysis} has more analyses.

%% file: 2_related_work.tex
\section{Related Work}\label{sec:related_work}

\textbf{Shortcut Issues in Machine Learning}
O.O.D. generalization~\citep{Krueger2021OutofDistributionGV,Hendrycks2020TheMF} has a long history of research in machine learning under names including ``learning under covariate shift''~\citep{Bickel2009DiscriminativeLU,Cao2011DistanceML}, ``simplicity bias''~\cite{Shan2020pitfalls,xu2019frequency,hu2020surprising}, ``anti-causal learning''~\citep{scholkopf2012causal}, and ``shortcuts''~\cite{geirhos2020shortcut}. 
Unlike prior solutions requiring collecting more data~\citep{de2019causal}, utilizing the biased model~\citep{clark2019dont,he2019unlearn}, obtaining extra labels~\citep{arjovsky2019invariant,ahuja2020invariant,Krueger2021OutofDistributionGV}, or making assumptions on the distribution~\citep{duchi2021learning,Delage2010DistributionallyRO,Sagawa2019DistributionallyRN}, we guide the optimization with additional domain knowledge about key inputs, and show  that this helps reduce shortcut issues. 

\textbf{Shortcuts in Image Classification.} Specific to image classification, many researchers have observed O.O.D. problems~\cite{bahng2020learning, he2021towards, arjovsky2019invariant}, and proposed to alleviate them by de-biasing~\citep{cadene2019rubi,bahng2020learning,zhang2021deep}, data augmentation~\citep{zhang2017mixup,devries2017improved,xu2021fourier} and causal intervention~\citep{arjovsky2019invariant,ahuja2020invariant,krueger2021out,wang2021causal} methods. 
In contrast, we propose to obtain the most task-relevant region in the image by unsupervised models to distract the optimization away from undesired shortcuts and show O.O.D.~generalization.

\textbf{Shortcuts in Behavior Cloning}
Behavior cloning (BC) suffers from O.O.D. issues because of the mismatch between the offline training distribution and online testing distribution~\citep{muller2006off,Ross2011,bansal2018chauffeurnet}.
In our experiments, we focus on resolving a specific aspect of distributional shift in BC from observation histories -- the ``copycat'' shortcut~\citep{chuan2020fighting}. Prior attempts to solve this include removing historical frames~\citep{muller2006off}, causal discovery~\citep{de2019causal}, history dropout~\citep{bansal2018chauffeurnet}, speed prediction regularization~\citep{codevilla2019exploring}, data re-weighting~\citep{chuan2021keframe}, and causal intervention~\citep{ortega2021shaking}. 
Instead, we focus on resolving the copycat shortcut by priming the policy with information from the most recent observation alone, achieving state-of-the-art performance.

%% file: 5_conclusion.tex
\section{Conclusion and Discussion}
\textbf{Summary:} In this paper, we propose PrimeNet, a simple and effective approach to resolve the shortcut issue in settings that permit domain knowledge of important inputs. PrimeNet is supported by recent theories on DNN training. On image classification and behavioral cloning tasks, our method outperforms the existing methods and significantly alleviates the shortcut issue to generalize beyond the training distribution. 

\textbf{Limitation:}
PrimeNet's gains come from the extra supervision in the form of correct/useful key input information. Though such domain knowledge is often easy to provide, it is indeed possible for poorly defined key inputs to hurt performance, which is a limitation of our method. Future work to address this, such as by automatically discovering key inputs from data, may make our method robust to such mis-specification.

%% file: 6_appendix.tex
\newpage
\appendix
\onecolumn

\section{The learned function in the toy experiment}
\label{app:toy-exp-curve}
\begin{figure*}[htb]
    \centering
    \includegraphics[width=\textwidth]{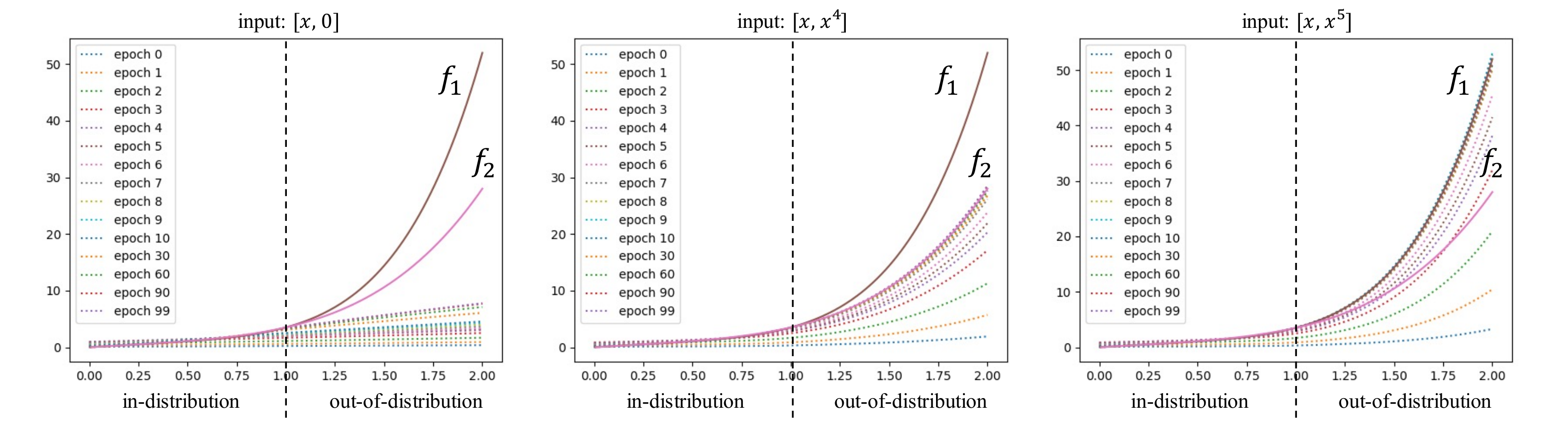}
    \caption{The curve of learned function of the neural network during training. The two solid lines are the ground truth curves of $f_{1}=1.5x^{5}+2x$ and $f_{2}=1.5x^{4}+2x$ respectively (we remove the additive Gaussian noise $\epsilon$ when plotting). The dashed lines are the functions learned by the MLP in different training epochs. We can see that if given different values of priming variable $\zeta$, the solution of the neural network will converge to different regions which are close to $\zeta$.}
    \label{fig:learned_func}
\end{figure*}

To verify how the priming variable $\zeta$ affects the final solution of the neural network in the toy regression experiment, we plot the curves of $\hat{y}$ vs. $x$ when providing different values of $\zeta$, 0 (no priming input) or $x^{4}$ or $x^{5}$. We plot the learned curves in different training epochs. As shown in Figure~\ref{fig:learned_func}, we can see that in the left subfigure, the model without priming input cannot learn an accurate solution and performs poorly in the out-of-distribution region. The middle and right subfigures show that if provided different priming variables, the neural networks will converge to different solutions. Furthermore, the final solution is close to $\zeta$, i.e. the priming variable guides the training of the NNs, which provides an empirical evidence for Proposition~\ref{prop: early}.

\section{Visualization results of NICO}
\label{app:nico-visualization}
\begin{figure}[h]
    \centering
    \includegraphics[width=\linewidth]{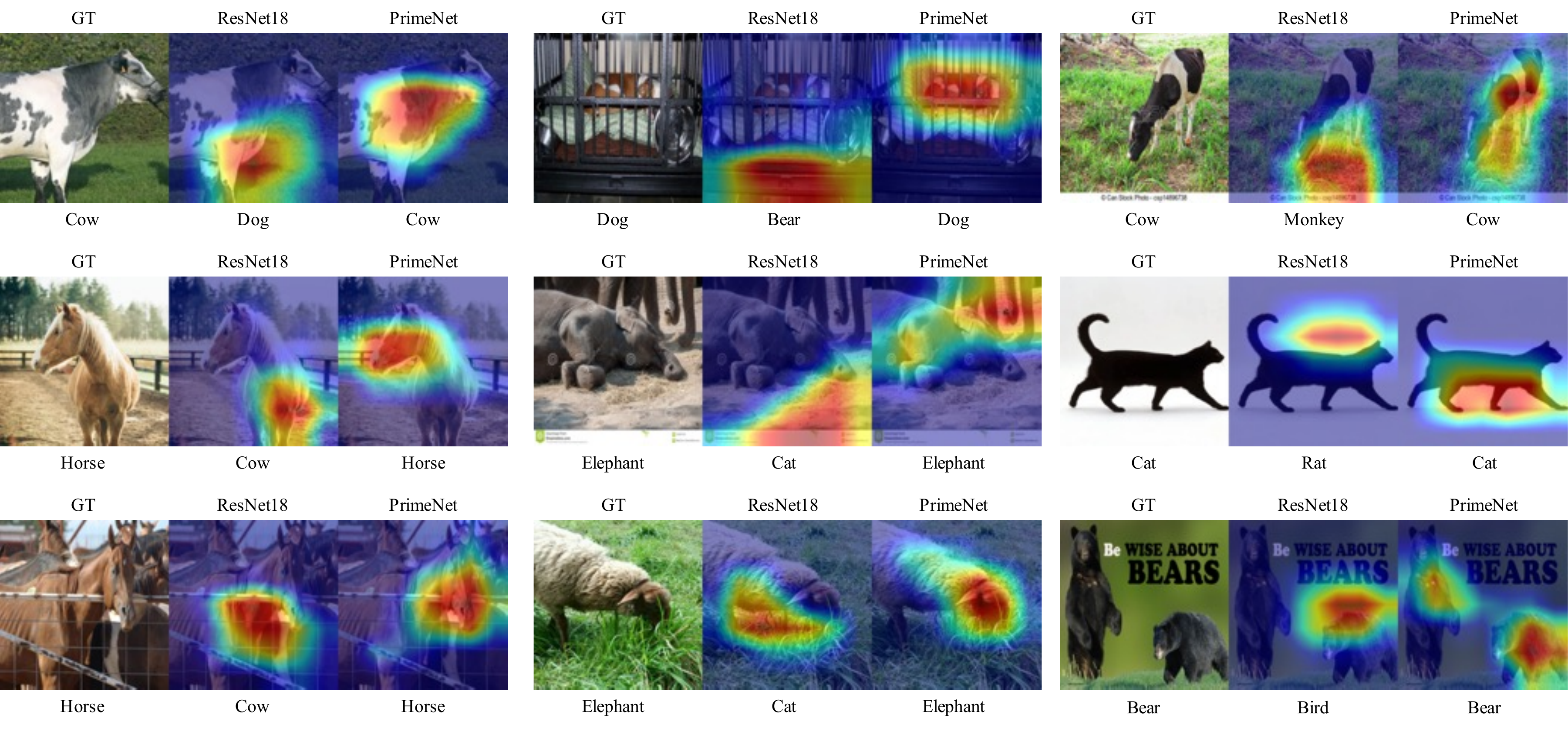}
    \caption{The activation maps of NICO classification model.}
    \label{fig:all-nico-grad-cam}
\end{figure}

Similar to Figure~\ref{fig:heatmap} for CARLA driving, Figure~\ref{fig:all-nico-grad-cam} shows activation maps for NICO. Vanilla ResNet18 learns a background shortcut and makes the wrong prediction, while PrimeNet correctly focuses on the discriminative foreground objects of the images.

\section{The detailed imitation results on CARLA Nocrash}
\label{app:more-carla-results}
The full results on Nocrash-Empty, Nocrash-Regular and Nocrash-Dense are shown in Table~\ref{tab:nocrash-empty-results}, Table~\ref{tab:nocrash-regular-results} and Table~\ref{tab:nocrash-dense-results}. We can see that our method performs significantly better than vanilla BC, key-input-only and other baselines. Our method gets highest \%success and lowest \#timeout, indicating that the shortcut in driving scenario, i.e. copycat problem, has been significantly alleviated.

\begin{table*}[htb]
\caption{CARLA \textit{Nocrash-Empty} results.}
\label{tab:nocrash-empty-results}
\begin{center}
\begin{small}
\begin{sc}
\begin{tabular}{c|ccccc}
\toprule
Method & \%success ($\uparrow$)  & \%progress ($\uparrow$)  & \#collision ($\downarrow$)   &  \#timeout ($\downarrow$)  \\
\midrule
vanilla bc    &  78.4 $\pm$ 11.6  &  83.2 $\pm$ 12.0  &  \textbf{1.6 $\pm$ 1.8}  &  20.0 $\pm$ 12.7  \\
key-input-only    &  44.9 $\pm$ 6.7  &  68.1 $\pm$ 5.3  &  39.7 $\pm$ 11.2  &  15.4 $\pm$ 5.4  \\
average-ensemble  &  84.1 $\pm$ 5.0  &  88.8 $\pm$ 2.8  &  1.0 $\pm$ 0.8  &  14.9 $\pm$ 5.3  \\
ours     &  \textbf{89.8 $\pm$ 1.4}  &  92.6 $\pm$ 0.8  &  1.6 $\pm$ 1.3  &  \textbf{8.6 $\pm$ 1.3}  \\
\midrule
fca      &  70.4 $\pm$ 7.6  &  86.1 $\pm$ 6.4  &  3.6 $\pm$ 1.8  &  26.0 $\pm$ 7.3  \\
Keyframe  &  \textbf{90.1 $\pm$ 5.7}  &  92.9 $\pm$ 2.3  &  \textbf{0.6 $\pm$ 0.7}  &  9.3 $\pm$ 5.4  \\
history-dropout  &  85.1 $\pm$ 3.7  &  \textbf{93.6 $\pm$ 2.6}  &  2.2 $\pm$ 1.3  &  12.7 $\pm$ 4.2  \\
dagger 150k    &  83.2 $\pm$ 9.4  &  90.4 $\pm$ 6.0  &  1.8 $\pm$ 0.9  &  15.0 $\pm$ 9.4  \\
\bottomrule
\end{tabular}
\end{sc}
\end{small}
\end{center}
\vskip -0.2in
\end{table*}

\begin{table*}[htb]
\caption{CARLA \textit{Nocrash-Regular} results.}
\label{tab:nocrash-regular-results}
\begin{center}
\begin{small}
\begin{sc}
\begin{tabular}{c|ccccc}
\toprule
Method & \%success ($\uparrow$)  & \%progress ($\uparrow$)  & \#collision ($\downarrow$)   &  \#timeout ($\downarrow$)  \\
\midrule
vanilla bc    &  67.1 $\pm$ 10.8  &  76.5 $\pm$ 10.2  &  \textbf{11.1 $\pm$ 3.1}  &  21.9 $\pm$ 12.7  \\
key-input-only    &  37.6 $\pm$ 6.1  &  61.2 $\pm$ 4.3  &  53.0 $\pm$ 7.9  &  10.2 $\pm$ 3.1  \\
average-ensemble  &  75.8 $\pm$ 3.6  &  84.4 $\pm$ 2.0  &  12.1 $\pm$ 3.0  &  12.4 $\pm$ 4.5  \\
ours     &  \textbf{81.1 $\pm$ 4.0}  &  \textbf{88.1 $\pm$ 2.6}  &  11.9 $\pm$ 3.3  &  \textbf{7.2 $\pm$ 2.8}  \\
\midrule
fca      &  58.0 $\pm$ 8.0  &  78.5 $\pm$ 7.1  &  14.7 $\pm$ 3.3  &  27.3 $\pm$ 8.8  \\
Keyframe  &  74.4 $\pm$ 7.3  &  83.2 $\pm$ 3.4  &  13.8 $\pm$ 2.7  &  11.9 $\pm$ 5.8  \\
history-dropout  &  70.1 $\pm$ 4.0  &  82.1 $\pm$ 2.2  &  18.3 $\pm$ 5.2  &  12.2 $\pm$ 4.4  \\
dagger 150k    &  69.7 $\pm$ 8.4  &  80.6 $\pm$ 6.0  &  14.8 $\pm$ 2.9  &  15.9 $\pm$ 8.5  \\
\bottomrule
\end{tabular}
\end{sc}
\end{small}
\end{center}
\vskip -0.2in
\end{table*}

\begin{table*}[htb]
\caption{CARLA \textit{Nocrash-Dense} results.}
\label{tab:nocrash-dense-results}
\begin{center}
\begin{small}
\begin{sc}
\begin{tabular}{c|ccccc}
\toprule
Method & \%success ($\uparrow$)  & \%progress ($\uparrow$)  & \#collision ($\downarrow$)   &  \#timeout ($\downarrow$)  \\
\midrule
vanilla bc    &  34.1 $\pm$ 7.5  &  62.2 $\pm$ 9.4  &  \textbf{30.2 $\pm$ 7.9}  &  36.1 $\pm$ 14.5  \\
key-input-only    &  13.1 $\pm$ 1.8  &  40.8 $\pm$ 3.0  &  76.4 $\pm$ 3.5  &  \textbf{11.1 $\pm$ 2.9}  \\
average-ensemble  &  41.7 $\pm$ 3.1  &  71.5 $\pm$ 3.2  &  43.7 $\pm$ 4.0  &  15.0 $\pm$ 0.8  \\
PrimeNet (ours)     &  \textbf{49.3 $\pm$ 3.6}  &  \textbf{75.0 $\pm$ 1.6}  &  39.4 $\pm$ 5.0  &  12.0 $\pm$ 1.9  \\
\midrule
fca~\citep{chuan2020fighting}      &  31.2 $\pm$ 5.2  &  66.5 $\pm$ 4.1  &  34.4 $\pm$ 8.1  &  35.3 $\pm$ 9.6  \\
Keyframe~\citep{chuan2021keframe}  &  41.9 $\pm$ 6.2  &  70.2 $\pm$ 4.0  &  33.9 $\pm$ 6.6  &  24.8 $\pm$ 7.9  \\
history-dropout~\citep{bansal2018chauffeurnet}  &  35.6 $\pm$ 3.5  &  67.0 $\pm$ 2.7  &  45.3 $\pm$ 3.5  &  20.3 $\pm$ 5.6  \\
dagger 150k~\citep{Ross2011}    &  42.7 $\pm$ 5.7  &  71.3 $\pm$ 1.9  &  35.0 $\pm$ 3.6  &  23.0 $\pm$ 7.1  \\
\bottomrule
\end{tabular}
\end{sc}
\end{small}
\end{center}
\vskip -0.2in
\end{table*}

\section{More ablation studies}
\label{app:more-ablation}
We conduct more ablation studies on CARLA and the detailed results are shown in Table~\ref{tab:carla-ablation}.

\begin{table*}[htb]
\caption{CARLA ablation study results.}
\label{tab:carla-ablation}
\begin{center}
\begin{small}
\begin{sc}
\begin{tabular}{c|ccccc}
\toprule
Method & \%success ($\uparrow$)  & \%progress ($\uparrow$)  & \#collision ($\downarrow$)   &  \#timeout ($\downarrow$)  \\
PrimeNet (ours)     &  49.3 $\pm$ 3.6  &  75.0 $\pm$ 1.6  &  39.4 $\pm$ 5.0  &  12.0 $\pm$ 1.9  \\
\midrule
early-fusion     &  51.4 $\pm$ 4.0  &  75.9 $\pm$ 2.1  &  37.4 $\pm$ 5.1  &  12.7 $\pm$ 2.6  \\
middle-fusion    &  49.6 $\pm$ 1.7  &  74.8 $\pm$ 1.9  &  37.3 $\pm$ 4.0  &  13.3 $\pm$ 3.0  \\
\midrule
resnet layer2 as $\zeta$    &  39.8 $\pm$ 4.0  &  69.4 $\pm$ 1.5  &  32.6 $\pm$ 1.5  &  28.1 $\pm$ 4.4  \\
resnet avg-pool as $\zeta$    &  44.8 $\pm$ 5.2  &  71.9 $\pm$ 4.7  &  40.0 $\pm$ 5.7  &  15.9 $\pm$ 3.4  \\
\midrule
$g_{\phi}(x)$ and $f_{\theta}(x,\zeta)$  &  37.1 $\pm$ 3.4  &  66.9 $\pm$ 2.8  &  34.1 $\pm$ 3.4  &  28.9 $\pm$ 6.0  \\
$g_{\phi}(x)$ and $f_{\theta}(k(x),\zeta)$  &  31.7 $\pm$ 5.7  &  62.2 $\pm$ 6.1  &  28.3 $\pm$ 9.0  &  40.0 $\pm$ 14.7  \\
$g_{\phi}(k(x))$ and $f_{\theta}(k(x),\zeta)$  &  13.8 $\pm$ 2.4  &  41.8 $\pm$ 4.5  &  74.9 $\pm$ 3.2  &  11.9 $\pm$ 2.6  \\
\midrule
w/o stop-gradient  &  44.4 $\pm$ 5.5  &  71.4 $\pm$ 3.0  &  39.2 $\pm$ 4.7  &  17.1 $\pm$ 2.1  \\
w/o end-to-end  &  47.3 $\pm$ 3.8  &   71.2 $\pm$ 1.2  &  43.4 $\pm$ 5.5  &  9.8 $\pm$ 3.1  \\
\bottomrule
\end{tabular}
\end{sc}
\end{small}
\end{center}
\vskip -0.2in
\end{table*}

\textbf{Different fusion stages.}
We can see that given the priming variable $\zeta$, it makes no difference where to inject it into $f_{\theta}$ (\%success ranges from $49.3$ to $51.4$ and other metrics are similar too), indicating that $\zeta$ provides a simpler shortcut than inferring and copying the previous action from the observation history. Wherever we inject $\zeta$, the neural network prefers to adopt it directly as the decision rather than expending greater effort to take the copycat shortcut. 

\textbf{Stop-gradient.}
Moreover, we find that the agent performs worse if we remove stop-gradient (\%success drops from 49.3 to 44.4), which is mainly due to timeout (increasing from 12.0 to 17.1). The increasing \#timeout shows that our method without stop-gradient still suffers from the shortcuts due to redundant information leakage during back-propagation.

\textbf{Two-stage training.}
And training $g_{\phi}$ and $f_{\theta}$ stage-by-stage, the performance also deteriorates (\%success drops from 49.3 to 47.3). We can see that its \#timeout is even fewer than ours but it gets a significantly higher \#collision (\#collision increases from 39.4 to 43.4), indicating that the agent trained stage-by-stage behaves more like key-input-only model, i.e. suffering from less copycat but failing to brake in time. We hypothesize that if we use a pretrained priming module $g_{\phi}$ at the beginning of $f_{\theta}$ training, $f_{\theta}$ will prefer to just copy the $\zeta$ as its output rather than learn from scratch to take accurate actions according to both the priming variable and observation history, which can also be viewed as overly relying on the "simpler" solution.

\section{More Analysis Experiments}
\label{app:more-analysis}
Then, because there are two inputs of main module $f_{\theta}$, we investigate the effect of the priming variable $\zeta$ and the input observation history $x$ on $f_{\theta}$.

\begin{figure*}[htb]
    \centering
    \includegraphics[width=0.75\textwidth]{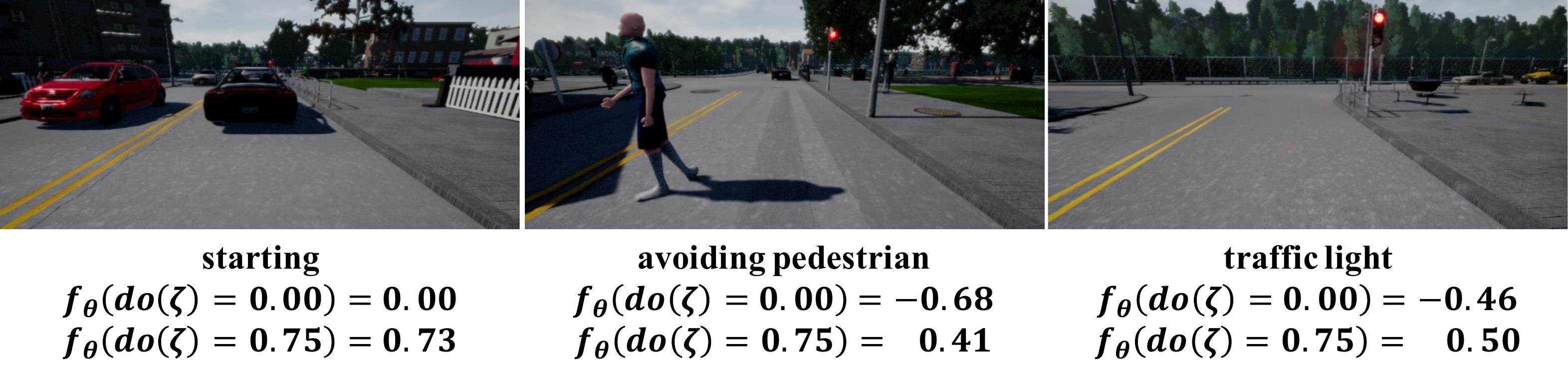}
    \caption{Examples of scenarios with high effects of the priming variable $\zeta$. The $f_{\theta}(do(\zeta)=0.00/0.75)$ means that we manually set the value of $\zeta$ to 0.00 and 0.75 respectively, fix the input $x$ and then get the output of the $f_{\theta}$. The larger the difference between $f_{\theta}(do(\zeta)=0.00)$ and $f_{\theta}(do(\zeta)=0.75)$, the stronger the effect of $\zeta$.}
    \label{fig:intervene_zeta}
\end{figure*}
\textbf{The effect of the priming variable $\zeta$.}
To study the effect of the priming variable in our model, we set the value of $\zeta$ manually while keeping all other factors the same. This process is similar to the intervention technique in causal inference literature~\citep{rubin1974estimating,rubin1978bayesian,pearl1995causal}, so we borrow the do-calculus $do(\zeta)$ from causal inference to denote this operation.
For easier interpretability, we only modify the acceleration dimension, i.e. setting throttle to 0 and 0.75 (the highest value in expert demonstrations), which can be denoted as $do(\zeta)=0/0.75$. The effect of $\zeta$ can be defined by $\text{Effect}=|f_{\theta}(do(\zeta)=0) - f_{\theta}(do(\zeta)=0.75)|$, where we omit another input variable $x$ for simplicity. Through studying the effect of $\zeta$ along the trajectories, we find that $\zeta$ tends to have a very high effect at some critical moments such as the examples in Figure~\ref{fig:intervene_zeta}, illustrating the important guidance effect of $\zeta$ in PrimeNet.
Especially, the effect of $\zeta_{t}$ is high when the car is starting, but it decreases after the car is started because at this time it is necessary to refine its actions according to observation history, which is what we expected.

\textbf{The effect of the input observation history $x$.}
Similarly, to study the effect of the input observation history $x$, we intervene it by repeating the current frame $H$ times, i.e. $do(x)=\left[o_{t},o_{t},\cdots\right]$ which creates counterfactual stationary cases, i.e. the previous action is 0. Recall the example in Section~\ref{sec:preliminaries}, to investigate what factors the agents use to determine whether to move forward or stop, we count the percentage of model outputs that change from accelerating to stop after we intervene on the input sequence, i.e. 
\begin{equation*}
    \frac{N(\text{speed}>0,f_{\theta}(x)>0,f_{\theta}(do(x))=0)}{N(\text{speed}>0,f_{\theta}(x)>0)}
\end{equation*}

\begin{figure}
  \begin{center}
    \includegraphics[width=0.4\linewidth]{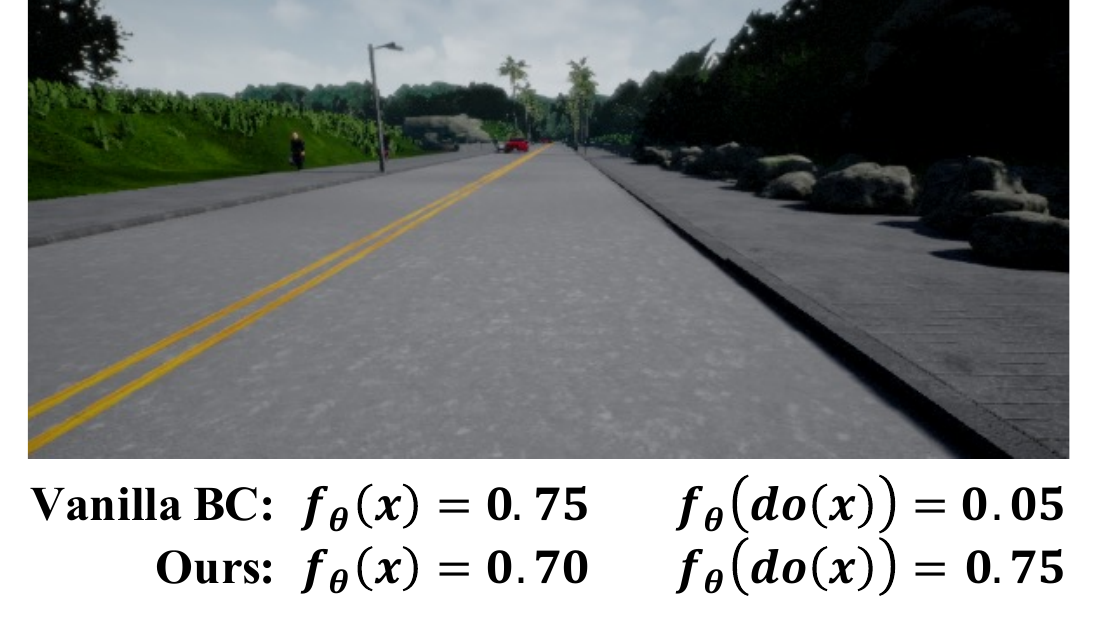}
  \end{center}
  \caption{An example of intervention on the input image sequence $x$.}
  \label{fig:intervene_obs}
\end{figure}
We count this metric for vanilla BC and our model on the same dataset. There are \textbf{66.43\%} samples changing from accelerating to stop in vanilla BC such as the example in Figure~\ref{fig:intervene_obs}, even though there is no signal to stop in the scene, e.g. vehicles, pedestrians, red lights or other obstacles. This illustrates that surprisingly in more than half of the cases, vanilla BC is making decisions only according to the previous action and ignores the current scene, which is causally incorrect. In the meanwhile, there is only \textbf{27.89\%} in our model, indicating that our model learns more correct causal relation and significantly alleviates the copycat shortcut.

\section{Baselines}
\label{app:baselines}
\textbf{Vanilla BC\&key-input-only.}
Vanilla BC and key-input-only baselines are naive behavioral cloning methods from observation history and the current observation respectively, which corresponds to the BC-OH and BC-SO in \citet{chuan2020fighting,chuan2021keframe}.

\textbf{Fighting-Copycat-Agents (FCA).}
\citet{chuan2020fighting} proposed to remove the unique information about the previous actions $a_{t-1:t-H}$ from the feature extracted from the observation history to prevent the agent to copy the $a_{t-1:t-H}$, based on adversarial learning.

\textbf{KeyFrame.}
\citet{chuan2021keframe} analyzed the copycat problem in terms of the imbalance data distribution and proposed a re-weighting method to up-weight the demonstration keyframes corresponding to expert action changepoints.

\textbf{History-Dropout.}
To address the copycat issue, \citet{bansal2018chauffeurnet} introduced a dropout on the observation history to randomly erase the channels of historical frames. We implement it by applying a Dropout layer~\citep{srivastava2014dropout} on the historical observations.

\textbf{Average-Ensemble.}
Average-Ensemble is a commonly used model combination approach. We implement it by averaging the outputs of vanilla BC and key-input-only at test time.

\textbf{DAGGER.}
DAGGER~\citep{Ross2011} is a widely used technique to address the distributional shift issue in behavioral cloning and is thought of as the \textbf{oracle} of imitation learning through online query.

\section{Additional Details on NICO Experiments}
\label{app:nico-exp-detail}

\begin{figure}
    \centering
    \includegraphics[width=0.5\textwidth]{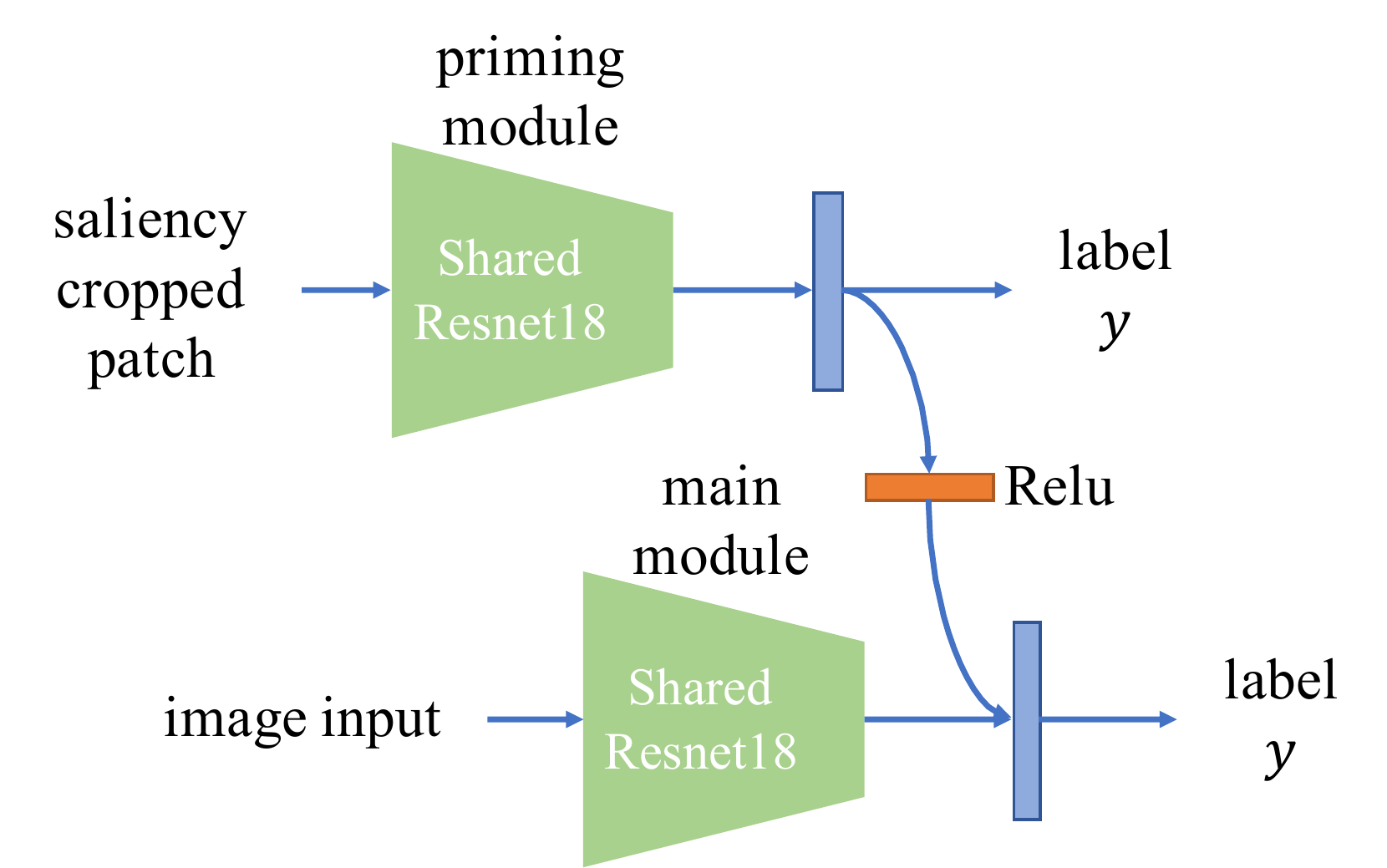}
    \caption{The architecture of our model for NICO experiments. We use a shared ResNet18 as the feature extractor. Use the output logits of priming module as $\zeta$ and feed it through a Relu layer to main module.}
    \label{fig:arch_our_nico}
\end{figure}

\textbf{Architectures}
As shown in Figure~\ref{fig:arch_our_nico}, there are two branches in our model. We use a share-weight ResNet18 as the feature extractor in both branches. For priming module, the input is the key input, i.e. an image patch cropped by saliency detection, and the output is a prediction about the label $y$. We use the output logits of priming module as priming variable $\zeta$. The logits are fed into a Relu layer and then concatenated with the feature of ResNet18 of main module. Next, the concatenated feature is put into the FC layer to make the final prediction.

\textbf{Training Details}
We use BASNet~\citep{qin2019basnet} as the key input extractor $k(\cdot)$, which is an unsupervised saliency detection model. We use Cross-Entropy loss and SGD optimizer to jointly train our model for 200 epochs. The initial learning rate is set to 0.05 and decays by 0.2 at 80, 120 and 160 epochs. We set the minibatch size to 128. We select the best hyper-parameters according to the validation accuracy and then test the models on the test set.

\section{Additional Details on CARLA Experiments}
\label{app:carla-exp-detail}
In Section~\ref{sec:experiments}, we briefly introduce the experiment setup of CARLA. More details are introduced below.

\textbf{Data Collection.}
The CARLA100 dataset~\citep{codevilla2019exploring} is collected by a PID controller. During collecting, 10\% expert actions are perturbed by noise~\citep{laskey2017dart}. We use three cameras: a forward-facing one and two lateral cameras facing 30 degrees away towards left or right~\citep{bojarski2016end}. Both noise injection and multiple cameras are common data augmentation techniques to alleviate distributional shift in autonomous driving.

\begin{figure}[htb]
    \centering
    \includegraphics[width=0.5\textwidth]{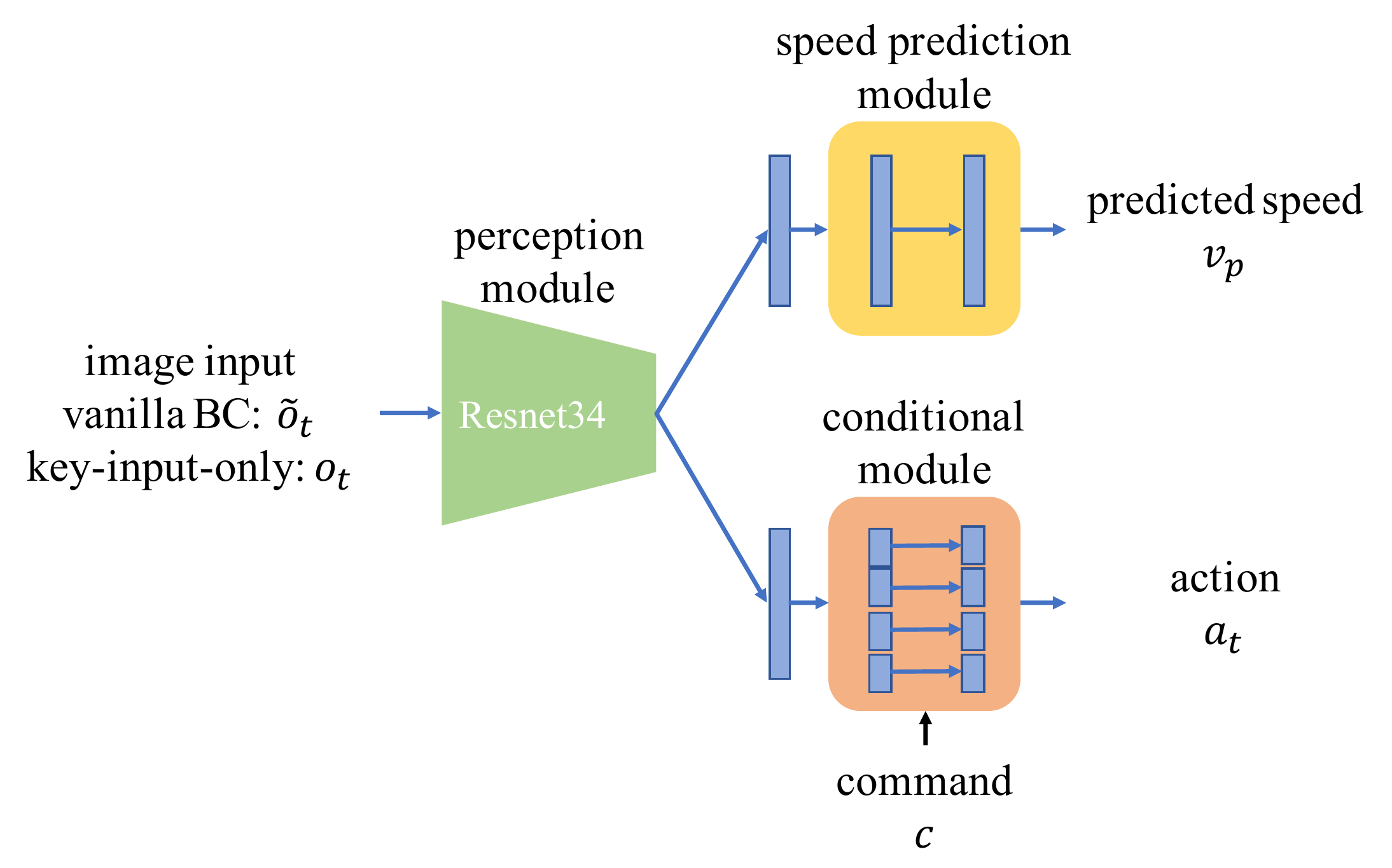}
    \caption{The conditional imitation learning architecture we used as our backbone. The input of vanilla BC is the observation history, denoted as $\tilde{o}_{t}$ for simplicity, and the input of key-input-only is the current observation $o_{t}$.}
    \label{fig:arch_bcso_bcoh}
\end{figure}
\begin{figure}[htb]
    \centering
    \includegraphics[height=4cm]{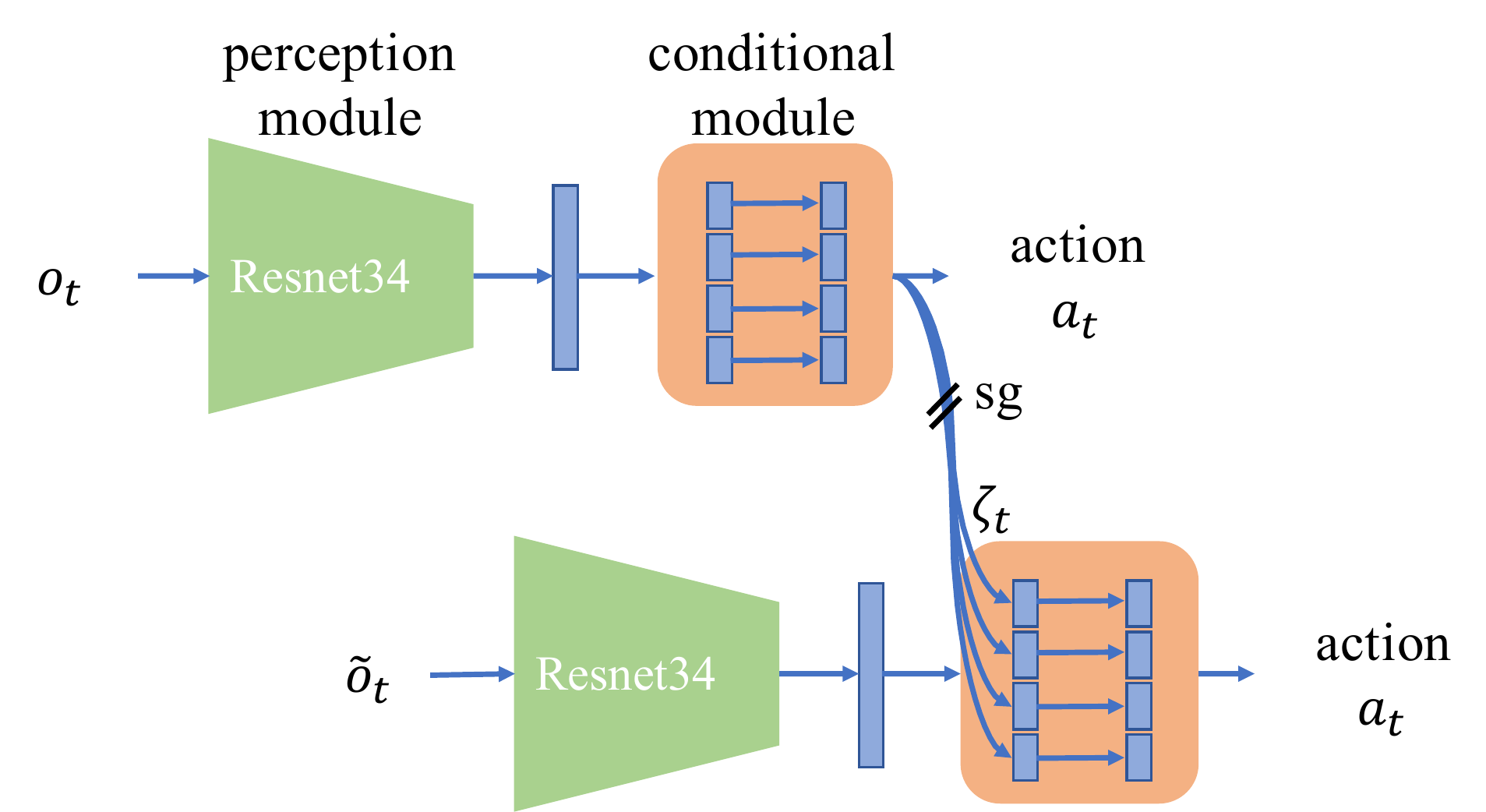}
    \caption{The architecture of our method. For simplicity, we omit the speed prediction module and the input command when drawing the figures, and sg means stop-gradient.}
    \label{fig:arch_ours}
\end{figure}
\begin{figure}[htb]
    \centering
    \subfigure[early fusion]{
    \includegraphics[height=3.2cm]{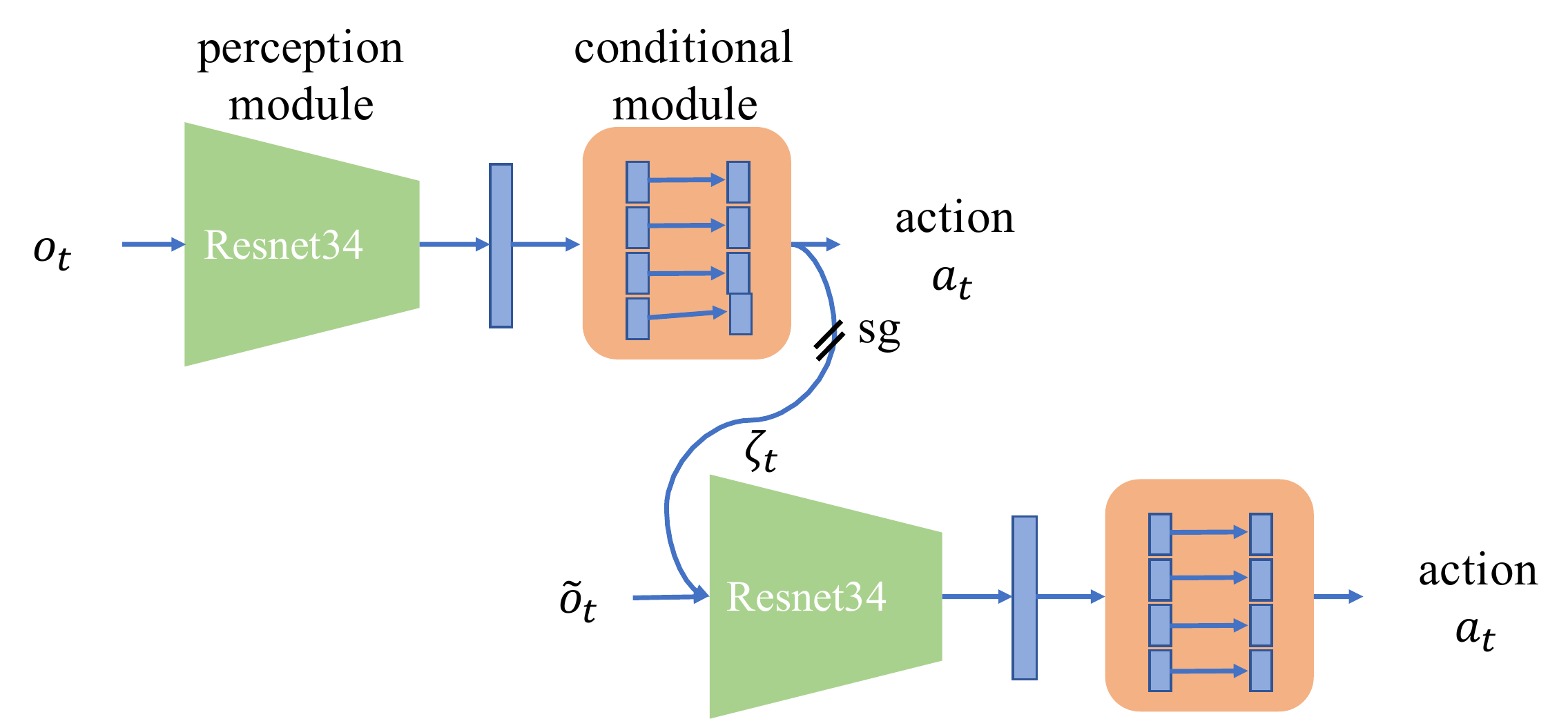}
    }
    \subfigure[middle fusion]{
    \includegraphics[height=3.2cm]{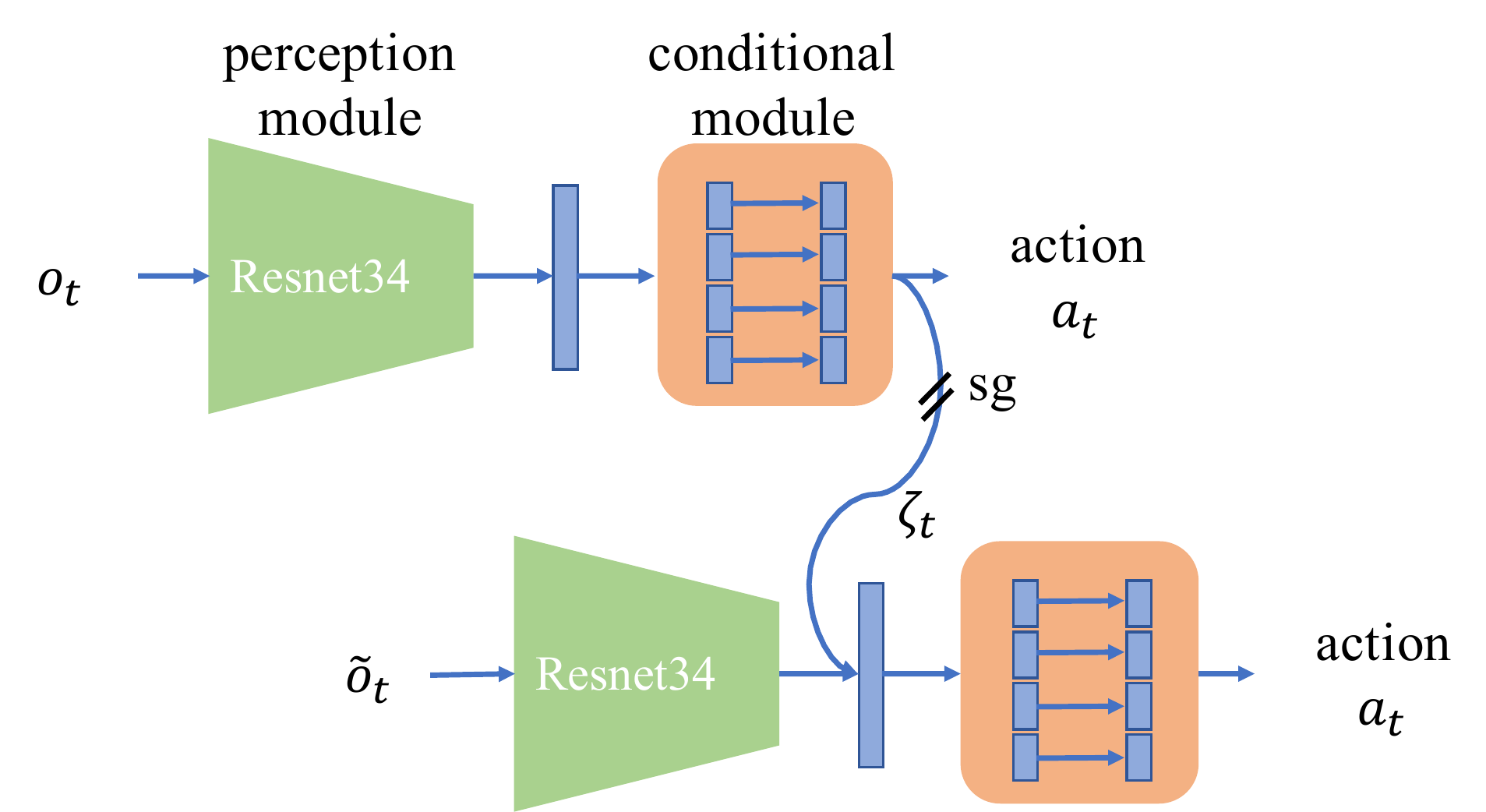}
    }
    \caption{The architectures of the early fusion model and middle fusion we study in Section~\ref{sec:ablation}.}
    \label{fig:arch_different_fusion}
\end{figure}

\textbf{Architectures.}
We use the backbone in conditional imitation learning framework CILRS~\citep{codevilla2019exploring} as our backbone. The only difference is that our model does not have the input speed (to create a pure POMDP~\citep{chuan2021keframe}). As shown in Figure~\ref{fig:arch_bcso_bcoh}, vanilla BC and key-input-only use the same architecture with different inputs, $\tilde{o}_{t}$ and $o_{t}$. Illustrated in Figure~\ref{fig:arch_ours}, our method integrates them together by concatenating the output of priming module with the features of the penultimate FC layer of main module. Moreover, the architectures of early fusion model and middle fusion model mentioned in Section\ref{sec:ablation} are shown in Figure~\ref{fig:arch_different_fusion}. In particular, early fusion means we concatenate the priming variable $\zeta$ with the input images, and middle fusion means that we concatenate it with the output feature of Resnet.

\textbf{Training Details.}
We use the $L_{1}$ loss to train all the models. We use Adam optimizer, set the initial learning rate to $2\times10^{-4}$ and decay the learning rate by 0.1 whenever the loss value no longer decreases for 5000 gradient steps. We set the minibatch size to 160 and train all the models until convergence (the learning rate equal to $1\times10^{-7}$).
Furthermore, we apply several commonly used techniques to our training process. We utilize the noise injection~\citep{laskey2017dart} and multi-camera data augmentation~\citep{bojarski2016end,giusti2015machine} to alleviate the distribution shift in offline imitation learning. All the models use the speed regularization~\citep{codevilla2019exploring} to address the copycat problem (also called inertia problem in their paper) to some extent. And we use ImageNet pretrained ResNet34~\citep{deng2009imagenet,he2016deep} as the perception module to get a better initialization~\citep{codevilla2019exploring} and the weighted control loss to balance the models' attention to each action dimension. Furthermore, different from the previous works~\citep{codevilla2018end,codevilla2019exploring,chuan2021keframe}, we use two-dimensional action space $a \in \left[-1, 1\right]^{2}$ for steering and acceleration, where the positive acceleration value means applying throttle and the negative one means braking. We empirically find that using acceleration as output modality is better than predicting throttle and brake separately among all baselines (except FCA) and our method.

\section{Additional Details on Mujoco Experiments}
\label{app:mujoco-exp-detail}
\textbf{Data Collection.} We first train an RL expert with PPO~\citep{schulman2017proximal} and use it to generate expert demonstration by rolling out in the environment. Specifically, we collect 10k samples for Ant and Walker2D, and 20k for Hopper based on imitation difficulty.

\begin{figure}[htb]
    \centering
    \includegraphics[height=4cm]{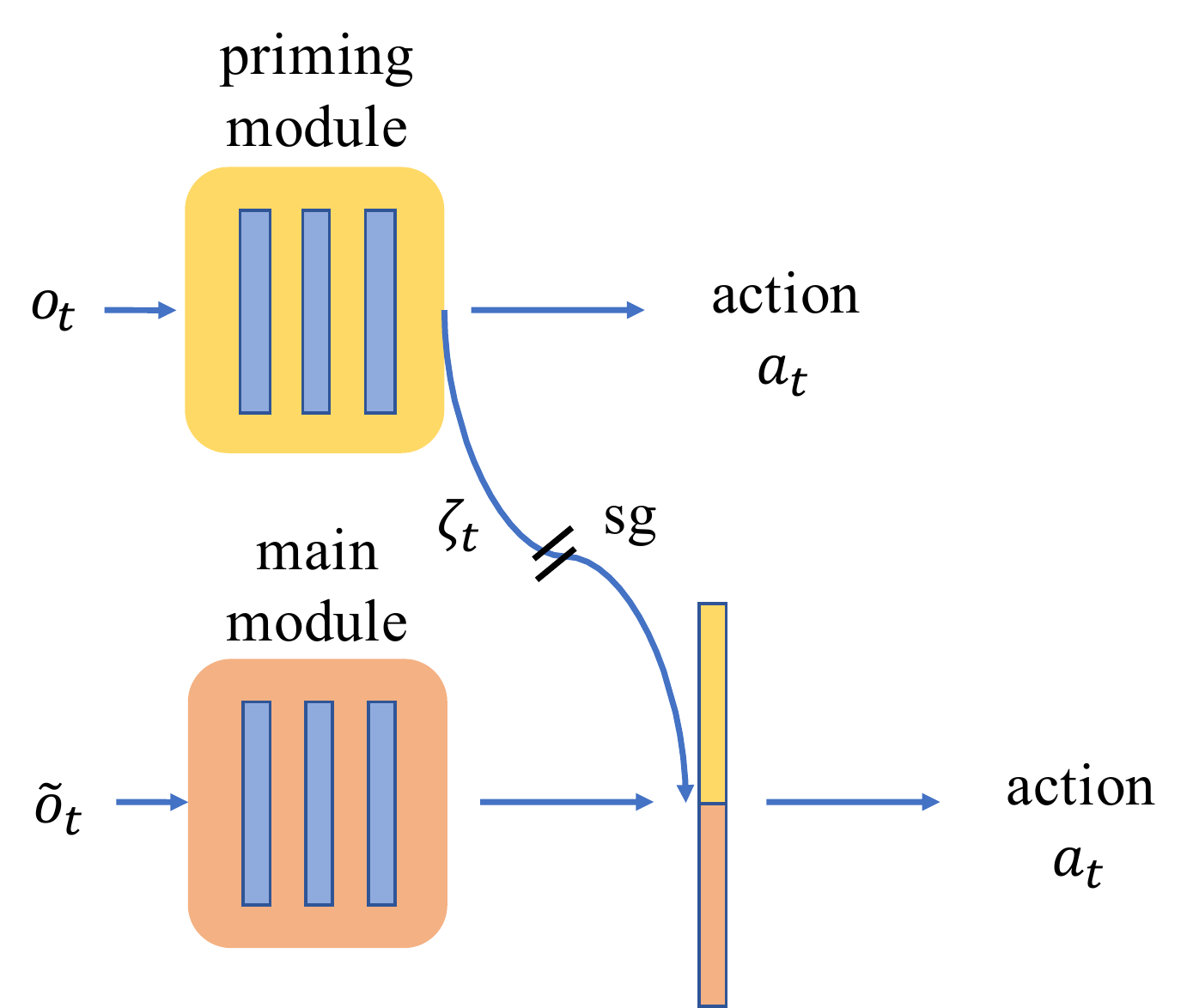}
    \caption{The architecture of our method for MuJoCo experiments, where sg means stop-gradient.}
    \label{fig:arch_mujoco}
\end{figure}

\textbf{Architectures.}
We follow a simple design for network architectures, shown in Figure~\ref{fig:arch_mujoco}. For both the priming module and main module, we use a three-layer MLP network. We concatenate the output action from the priming module to the output of main module, and input that to another fully-connected layer for the final output action.

\textbf{Training Details.}
We use MSE loss and Adam optimizer to train all models. We use a learning rate of 1e-4 for both HalfCheetah and Ant, and 1e-5 for Hopper and use linear learning rate decay. For each environment, we train it for 1000 epochs until convergence. We set the minibatch size to 64. We train each method three times, and report the mean and standard deviation of evaluation rewards for the last three evaluation steps.

\input{6_appendix_theory}

%% file: 6_appendix_theory.tex
\section{Theorem Version for Proposition~\ref{prop: early}}
\label{app:theoremFormal}

We next formally state our theorem, starting from the basic definitions and assumptions.

\subsection{Theorem Statement}

\textbf{Data Distribution.}
Let $(\x, \y) \sim \cP_{in} \subset \bR^d \times \bR$, $(\z, \y) \sim \cP_{out} \subset \bR^d \times \bR$ denote the input and the corresponding label in the \emph{training region} and \emph{test region}, respectively.
Without loss of generality, we assume that $\bE \x =  \bE \z = \boldsymbol{0}$, and $ |\y| \leq O(1)$ is bounded almost surely.
For simplicity, we assume the data covariance satisfies $\Sigma = \Sigma_{\x} = \Sigma_{\z}$, and assume that $\operatorname{Tr}(\Sigma) = d$, $\sigma_{min} (\Sigma) = O(1)$, and $\sigma_{max} (\Sigma) = O(1)$.
We assume that the re-scaled input $\Sigma^{-1/2} \x$ and $\Sigma^{-1/2} \z$ has independent and $O(1)$-subGaussian coordinates.

Denote the optimal linear estimator as $\beta^*$, which satisfies $\bE_{\cP_{in}}(\y - \x^\top \beta^*)^2 = \min_\beta \bE_{\cP_{in}}(\y - \x^\top \beta)^2$, and $\|\beta^* \| = O(1)$.
We assume that the residual $\boldsymbol{\epsilon} = \y - \x^\top \beta^*$ is mean zero, independent, $O(1)$-subgaussian conditional on $\x$ with $\bE \boldsymbol{\epsilon}^2|\x = O(1) $.
Given the dataset with $n$ pairs independently sampled from $\cP_{in}$ and $\cP_{out}$, we denote the empirical distribution by $\hat{\cP}_1^n$ and $\hat{\cP}_2^n$.
We assume that the sample size satisfies $n \gg d$ and $n = d^{O(1)}$, and the corresponding design matrix is diagonalizable.

\textbf{Model.}
We consider a two-layer fully-connected neural network with $m$ hidden neurons: $f(\x; \w, v):=\frac{1}{\sqrt{m}} \sum_{r=1}^m v_r \sigma(\w_r^\top\x/\sqrt{d})$ with symmetric initialization, where the activation $\sigma$ is smooth or piece-wise linear.
In this paper, \emph{Neural Tangent Kernels (NTK)} are applied to approximate the neural networks\footnote{The neural tangent kernel regime is widely considered in theoretical analysis as a surrogate to neural networks, since neural networks converge to the neuron tangent kernel as its width goes to infinity.}, where $\{ \w_r \}$ is trainable and $\{v_r\}$ is fixed following the regime in \citet{DBLP:conf/icml/AroraDHLW19}.
For the training process with constant step size (sufficiently small) and $\ell_2$ loss, we derive the following Theorem~\ref{thm: mainTheorem} for the function $f_{NTK}(\cdot)$ trained at time $t$

For the neural networks, we consider a two-layer fully connect one and use the matrix formulation $(\W, \vb)$ as follows:
\begin{equation*}
    f(\x; \W, \vb)=\frac{1}{\sqrt{m}} \sum_{r=1}^m v_r \sigma(\w_r^\top\x/\sqrt{d}) = \frac{1}{\sqrt{m}}\vb^\top \sigma(\W\x/\sqrt{d}).
\end{equation*}

\textbf{Linear Kernels.} Follong~\citet{DBLP:conf/nips/HuXAP20}, when considering linear models, we consider the linear feature
On the other hand, we consider the linear feature
\begin{equation*}
    \phi_{LIN}(X) = \frac{1}{\sqrt{d}} \left[ \begin{array}{c}
     \zeta \x \\
    \nu 
\end{array}\right],
\end{equation*}
where $\zeta = \bE[\sigma^\prime(g)]$ and $\nu = \bE[g\sigma^\prime(g)] \cdot \sqrt{\operatorname{Tr}[\Sigma^2]/d}$.
We denote the function trained using linear kernels at time $t$ as $f_{LIN}^{(t)}$, and define the corresponding predictions on $\X$ and $\Z$ as $\hat{Y}_{LIN}(\X)$ and $\hat{Y}_{LIN}(\Z)$. 

\textbf{Neural Tangent Kernels.} We consider a neural tangent kernel using the following feature
$$\phi_{NTK}(X) = \operatorname{Vec} \left(\frac{\partial f(\x; \W, \vb)}{\partial \W}\right).$$
We denote the function trained using neural tangent kernels at time $t$ as $f_{NTK}^{(t)}$, and define the corresponding predictions on $\X$ and $\Z$ as $\hat{Y}_{NTK}(\X)$ and $\hat{Y}_{NTK}(\Z)$. 

\textbf{Symmetric Initialization} In this paper, we apply the symmetric initialization, $\boldsymbol{w}_i \sim \cN(\boldsymbol{0}, \boldsymbol{I}_d)$ and $v_i \sim \operatorname{Unif}(\{1, -1\})$, $\boldsymbol{w}_{(i+m/2)} = \boldsymbol{w}_i$, $v_{(i+m/2)} = -v_i$, $i = 1 \dots, m/2$.
This type of symmetric initializationfollowing \citep{DBLP:conf/nips/ChizatOB19,DBLP:conf/msml/ZhangXLM20,DBLP:conf/iclr/HuLY20,DBLP:conf/iclr/BaiL20,DBLP:conf/nips/HuXAP20} guarantees that $f(\x; \W, \vb) = 0$ at initialization.
Therefore, we can regard the corresponding NTK is trained by starting with initialization $\boldsymbol{0}$.

\subsection{The Main Theorem}
We now provide the main theorem in Theorem~\ref{thm: mainTheorem}

\begin{theorem}
\label{thm: mainTheorem}
Let $\alpha \in (0, 1/4)$ be a fixed constant. 
Let $h(\x) = \x^\top \beta^*: \bR^d \to \bR$ denote a linear model, and $s: \bR^d \to \bR$ denote another model.
We assume the following (A1-A3) assumptions\footnote{
We denote $ \| x_1 - x_2 \|_{\mu} = (\bE_{\mu} (x_1 - x_2)^2 )^{1/2}$}:
\begin{enumerate}
    \item[A1] In the training region $\cP_{in}$, function $f$ and $s$ are both close to the ground truth, 
    $\| y -  h(\x) \|_{\cP_{in}} \leq \epsilon$, and $\| y -  s(\x) \|_{\cP_{in}} \leq \epsilon$.
    \item[A2] In the test region $\cP_{out}$, function $f$ and $s$ are separable,  $\| s(\z) -  h(\z) \|_{\cP_{out}}\geq O(1)$.
    \item[A3] The width of the neural network satisfies $m =\Omega(  d^{1+\alpha} )$ 
\end{enumerate}
Then for the model $f_{NTK}^{t}(\cdot)$ trained by neural tangent kernel at time $t = \Theta(d^{1+\frac{1}{3}\alpha})$, satisfies the following statements (C1-C2) hold with high probability\footnote{The probability is taken over the random initialization, the training samples, and the test samples}:
\begin{enumerate}
    \item[C1] In training region $\cP_{in}$, the trained model reaches small training error, $\| f_{NTK}(\x) - y \|_{\hat{\cP}_1^n} - \epsilon \lesssim d^{-\frac{1}{3} \alpha} + \sqrt{\frac{d}{n}}$.
    \item[C2] In test region $\cP_{out}$, the trained model is closer to $f$ instead of $s$, $\| f_{NTK}(\z) - h \|_{\hat{\cP}_2^n} \lesssim d^{-\frac{1}{3} \alpha} + \sqrt{\frac{d}{n}}$ and $\| f_{NTK}(\z) - s \|_{\hat{\cP}_2^n} = \Omega(1)$. 
\end{enumerate}
\end{theorem}

\subsection{Proof}
In this section, We provide the whole proof of Theorem~\ref{thm: mainTheorem}.
\subsection{Proof}
\begin{proof}
We combine the following Three Lemmas to reach the final conclusion of Theorem~\ref{thm: mainTheorem}.
We first show in Lemma~\ref{lem: trainRegion} that the model trained by NTK is indeed close to the model trained by linear predictions in the train region $\cP_{in}$.
We then show in Lemma~\ref{lem: testRegion} that the trained model is also close to the model trained by linear predictions in the test region $\cP_{out}$.
We finally show in Lemma~\ref{lem: linear training} that linear models can approximate the ground truth function for a proper time $t$.
\end{proof}

\begin{lemma}[Bounding difference between NTK and linear predictions (train region)]
\label{lem: trainRegion}
Under the settings in Theorem~\ref{thm: mainTheorem}, the prediction of the NTK models ($\hat{Y}_{NTK}(\X)$) is close to the prediction of linear models ($\hat{Y}_{LIN}(\X)$), namely,
\begin{equation*}
        \frac{1}{\sqrt{n}} \left\| \hat{Y}_{NTK}(\X) - \hat{Y}_{LIN}(\X) \right\|
    \lesssim  d^{-\frac{1}{3} \alpha}.
\end{equation*}
\end{lemma}

\begin{proof}[Proof of Lemma~\ref{lem: trainRegion}]
Note that NKT regimes is overparameterized while linear regime is underparameterized.
We derive from Proposition~\ref{prop: Trajectory Analysis} that the prediction (on training set) from neural tangent kernel and the linear kernel is
\begin{equation*}
    \begin{split}
        &\hat{Y}_{NTK}(\X) = \left[I - \left[I - \frac{\lambda}{n}\phi_{NTK}(X) \phi_{NTK}^\top(X)\right]^t \right]Y,\\
        &\hat{Y}_{LIN}(\X) = \left[I - \left[I - \frac{\lambda}{n}\phi_{LIN}(X) \phi_{LIN}^\top(X)\right]^t \right]Y.
    \end{split}
\end{equation*}

Note that function $(1 - \lambda x)^t$ is $\lambda t$-Lipschitz, we have
\begin{equation*}
\begin{split}
    &\frac{1}{\sqrt{n}} \left\| \hat{Y}_{NTK}(\X) - \hat{Y}_{LIN}(\X) \right\| \\
    =&\frac{1}{\sqrt{n}} \left\| \left[\left[I - \frac{\lambda}{n}\phi_{LIN}(X) \phi_{LIN}^\top(X)\right]^t- \left[I -\frac{\lambda}{n}\phi_{NTK}(X) \phi_{NTK}^\top(X)\right]^t \right]Y \right\|, \\
    \leq & \frac{1}{\sqrt{n}} \left\| \left[\left[I - \frac{\lambda}{n}\phi_{LIN}(X) \phi_{LIN}^\top(X)\right]^t- \left[I -\frac{\lambda}{n}\phi_{NTK}(X) \phi_{NTK}^\top(X)\right]^t \right]\right\| \left\| Y \right\| \\
    \overset{(i)}{\leq}  & \frac{1}{\sqrt{n}} \lambda t \frac{1}{n} \log(n) \left\| \phi_{LIN}(X) \phi_{LIN}^\top(X) - \phi_{NTK}(X) \phi_{NTK}^\top(X) \right\| \left\| Y \right\| \\
    \overset{(ii)}{=}& \frac{1}{\sqrt{n}} \lambda t \frac{1}{n} \log(n) \cdot O\left(\frac{n}{d^{1+\alpha}}\right) \cdot O(\sqrt{n}) \\
    =& O\left(\lambda t \log(n) \frac{1}{d^{1+\alpha}} \right),
\end{split}
\end{equation*}
where we apply Proposition~\ref{prop: matrix function} in (i) and apply Proposition~\ref{prop: bounding train kernels} in (ii). 
And we use the fact that $\y$ is bounded almost surely to get $\|Y\| \leq O(\sqrt{n})$. 
Due to the fact that $n = d^{O(1)} $ and $t = \Theta\left( d^{1+\frac{1}{3}\alpha}\right)$, we have
\begin{equation*}
\begin{split}
    \frac{1}{\sqrt{n}} \left\| \hat{Y}_{NTK}(\X) - \hat{Y}_{LIN}(\X) \right\|
    \lesssim  d^{-\frac{1}{3} \alpha}.
\end{split}
\end{equation*}

\end{proof}

\begin{lemma}[Bounding difference between NTK and linear predictions (test region)]
\label{lem: testRegion}
Denote $\Z \in \bR^{n \times p}$ the test samples independent sampled from distribution $\cP_{out}$.
Under the settings in Theorem~\ref{thm: mainTheorem}, the prediction of the NTK models ($\hat{Y}_{NTK}(\Z)$) is close to the prediction of linear models ($\hat{Y}_{LIN}(\Z)$) at test region with high probability, namely,
\begin{equation*}
        \frac{1}{\sqrt{n}} \left\| \hat{Y}_{NTK}(\Z) - \hat{Y}_{LIN}(\Z) \right\|
    \lesssim d^{-\frac{1}{3} \alpha}.
\end{equation*}
\end{lemma}

\begin{proof}[Proof of Lemma~\ref{lem: testRegion}.]
Similar to the proof in Lemma~\ref{lem: trainRegion}, the predictions are

\begin{equation*}
    \begin{split}
        &\hat{Y}_{NTK}(\Z) \\
        =& \phi_{NTK}(Z) \phi_{NTK}^\top(X) \left[I - \left[I - \frac{\lambda}{n}\phi_{NTK}(X) \phi_{NTK}^\top(X)\right]^t \right] \left[ \phi_{NTK}(X) \phi_{NTK}^\top(X) \right]^{-1} Y,\\
        &\hat{Y}_{LIN}(\Z) \\
        =& \phi_{LIN}(Z) \phi_{LIN}^\top(X) \left[I - \left[I - \frac{\lambda}{n}\phi_{LIN}(X) \phi_{LIN}^\top(X)\right]^t \right] \phi_{LIN}(X) \left[\phi_{LIN}(X) \phi_{LIN}^\top(X) \right]^{-2} \phi_{LIN}^{\top}(X) Y.
    \end{split}
\end{equation*}

We calculate that
\begin{equation*}
    \begin{split}
        &\hat{Y}_{NTK}(\Z)  - \hat{Y}_{LIN}(\Z) \\
        =& \phi_{NTK}(Z) \phi_{NTK}^\top(X) \left[I - \left[I - \frac{\lambda}{n}\phi_{NTK}(X) \phi_{NTK}^\top(X)\right]^t \right] \left[ \phi_{NTK}(X) \phi_{NTK}^\top(X) \right]^{-1} Y\\
        &- \phi_{LIN}(Z) \phi_{LIN}^\top(X) \left[I - \left[I - \frac{\lambda}{n}\phi_{LIN}(X) \phi_{LIN}^\top(X)\right]^t \right] \phi_{LIN}(X) \left[\phi_{LIN}(X) \phi_{LIN}^\top(X) \right]^{-2} \phi_{LIN}^{\top}(X) Y\\
        =& \underline{\frac{\lambda}{n} \left[\phi_{NTK}(Z) \phi_{NTK}^\top(X) - \phi_{LIN}(Z) \phi_{LIN}^\top(X) \right]  A_1 Y} + \underline{ \frac{\lambda}{n} \phi_{LIN}(Z) \phi_{LIN}^\top(X) [A_1 - A_2] Y}\\
        \triangleq& \text{\textcircled{1}}  + \text{\textcircled{2}},
    \end{split}
\end{equation*}
where $A_1 = \left[I - \left[I - \frac{\lambda}{n}\phi_{NTK}(X) \phi_{NTK}^\top(X)\right]^t \right] \left[ \frac{\lambda}{n} \phi_{NTK}(X) \phi_{NTK}^\top(X) \right]^{-1}$, and\\ $A_2 = \left[I - \left[I - \frac{\lambda}{n}\phi_{LIN}(X) \phi_{LIN}^\top(X)\right]^t \right] \phi_{LIN}(X) \left[\phi_{LIN}(X) \phi_{LIN}^\top(X) \right]^{-2} \phi_{LIN}^{\top}(X)$.

\textbf{Bounding \textcircled{1}.}
We derive from Proposition~\ref{prop: bounding cross kernels} that 
$$\left\| \phi_{NTK}(Z) \phi_{NTK}^\top(X)  - \phi_{LIN}(Z) \phi_{LIN}^\top(X)  \right\| \lesssim \frac{n}{d^{1+\alpha}}.$$

We then calculate that the eigenvalues of the matrix $A_1$ can be bounded by
$$
\frac{1 - (1 - \lambda \sigma_i^t)}{\lambda \sigma_i} \leq t,
$$
where $\sigma_i$ denotes the eigenvalues of matrix $\phi_{NTK}(Z) \phi_{NTK}^\top(X)$.
Note that $\| \Y \| = O(\sqrt{n})$ due to the bounded assumption.
Therefore, we have
\begin{equation}
    \begin{split}
        &\text{\textcircled{1}} \\
        \leq & \frac{\lambda}{n}   \left\|\phi_{NTK}(Z) \phi_{NTK}^\top(X) - \phi_{LIN}(Z)  \phi_{LIN}^\top(X) \right\| \cdot \| A_1\| \cdot \| \Y \| \\
        \lesssim & \frac{\lambda}{n} \cdot \frac{n}{d^{1+\alpha}} \cdot t \cdot \sqrt{n} \\
        \lesssim & \left(\frac{ t \sqrt{n}}{d^{1+\alpha}}\right).
    \end{split}
\end{equation}

\textbf{Bounding \textcircled{2}.}
We first use Proposition~\ref{prop: matrix function} to bound $\|A_1 - A_2 \|$.
The eigenvalues of $A_1$ and $A_2$ has the form of $(1 - (1-\lambda \sigma_i)^t)/\lambda \sigma_i$ where $\sigma_i$ are the eigenvalues of $\phi_{NTK}(X) \phi_{NTK}^\top(X)$ and  $phi_{LIN}(X) \phi_{LIN}^\top(X)$.
From Proposition~\ref{prop: ex function Lip}, we see that the Lipschitz constant $L_e$ for function $e(x) = (1 - (1-x)^t)/x$ satisfies $L_e \lesssim t^2$.
Therefore, 
\begin{equation*}
    \begin{split}
        \| A_1 - A_2\| &\lesssim t^2 \log(n) \|\frac{\lambda}{n} \phi_{NTK}(X) \phi_{NTK}^\top(X) - \frac{\lambda}{n} \phi_{LIN}(X) \phi_{LIN}^\top(X) \| \\
        &\lesssim t^2 \cdot \frac{\lambda}{n} \cdot \frac{n}{d^{1+\alpha}} \\
        &\lesssim t^2 \frac{\lambda}{d^{1+\alpha}}
    \end{split}
\end{equation*}

Besides, we have that $\phi_{LIN}(Z) \phi_{LIN}^\top(X) \leq \|\phi_{LIN}(Z) \phi_{LIN}^\top(X) - \phi_{NTK}(Z) \phi_{NTK}^\top(X) \| + \|\phi_{NTK}(Z) \phi_{NTK}^\top(X) \| \lesssim n/d$, and therefore,
\begin{equation}
    \begin{split}
        &\text{\textcircled{2}} \\
        =& \frac{\lambda}{n} \phi_{LIN}(Z) \phi_{LIN}^\top(X) [A_1 - A_2] \Y\\
        \leq & \frac{\lambda}{n} \| \phi_{LIN}(Z) \phi_{LIN}^\top(X) \| \cdot \| A_1 - A_2 \| \cdot \| \Y\| \\
        \leq & \frac{\lambda}{n} \cdot O\left(\frac{n}{d}\right) \cdot O\left(t^2 \frac{\lambda}{d^{1+\alpha}} \right) \cdot O(\sqrt{n})\\
        \lesssim & \frac{t^2 \sqrt{n}}{d^{2+\alpha}}. 
    \end{split}
\end{equation}

Therefore, we have that
\begin{equation*}
\begin{split}
    \frac{1}{\sqrt{n}} \| \hat{Y}_{NTK}(\Z)  - \hat{Y}_{LIN}(\Z)  \| &\lesssim \frac{ t }{d^{1+\alpha}} + \frac{t^2}{d^{2+\alpha}} \\
    &\lesssim d^{- \frac{1}{3} \alpha}.
\end{split}
\end{equation*}
where the last equation is due to $t = \Theta(d ^{1+\frac{1}{3}\alpha})$.
\end{proof}

\begin{lemma}[Linear Training]
\label{lem: linear training}
Denote $ \hat{\Y}_{LIN}$ as the predictions of the linear predictors, $\Y(\X)$ as the true label, and $\hat{\Y}^*_{LIN}$ as the best linear predictor. 
Under the settings in Theorem~\ref{thm: mainTheorem}, the predictions of $f_{LIN}$ is close to the ground truth in both $\cP_{in}$ and $\cP_{out}$. That is to say: with high probability
\begin{equation*}
\begin{split}
    &\frac{1}{\sqrt{n}} \| \hat{Y}_{LIN}(\X)  - \Y(\X) \|_2 \lesssim \sqrt{\frac{d}{n}} + \epsilon, \\
     &   \frac{1}{\sqrt{n}} \| \hat{Y}_{LIN}(\Z)  - \hat{Y}_{LIN}^*(\Z) \|_2 \lesssim \sqrt{\frac{d}{n}} .
\end{split}
\end{equation*}

\end{lemma}

\begin{proof}[Proof of Lemma~\ref{lem: linear training}]
It suffices to bound the parameters trained by $\phi_{LIN}(\X)$ with the optimal parameter $\bar{\beta}^{*}$.
Note that ${\beta}_{LIN}^{*}$ is a re-scaled version of $\beta^*$, since $\phi_{LIN}(\X)$ has a $1/\sqrt{d}$ scale.
Therefore, informally, $\phi_{LIN}^\top(\X) \phi_{LIN}(\X) \approx \frac{1}{d} \X^\top \X$ and $\beta_{LIN}^* \approx \sqrt{d} \beta^*$.

We bound the difference of $\Y$ from the parameter perspective. 
Note that by gradient descent starting from initialization starting from initialization $\boldsymbol{0}$, by Proposition~\ref{prop: Trajectory Analysis}, we have
\begin{equation*}
    \begin{split}
        &\beta_{LIN}^{(t)} = \left[ I - \left[I - \frac{\lambda}{n} \phi_{LIN}^\top(\X_1) \phi_{LIN}(\X_1)\right]^t \right] \left[\phi_{LIN}^\top(\X_1) \phi_{LIN}(\X_1)\right]^{-1} \phi_{LIN}^\top(\X_1) \Y.
    \end{split}
\end{equation*}

By plugging into $\Y = \X^\top \beta^* + \boldsymbol{\epsilon} = \phi_{LIN}(\X) \beta_{LIN}^* +\boldsymbol{\epsilon}$ where $\boldsymbol{\epsilon} = \Y - \X^\top \beta$, it holds that
\begin{equation*}
    \begin{split}
    \beta_{LIN}^{(t)} =&[ I - [I - \frac{\lambda}{n} \phi_{LIN}^\top(\X_1) \phi_{LIN}(\X_1)]^t ]\beta_{LIN}^* \\
    &+ [ I - [I - \frac{\lambda}{n} \phi_{LIN}^\top(\X_1) \phi_{LIN}(\X_1)]^t ] [\phi_{LIN}^\top(\X_1) \phi_{LIN}(\X_1)]^{-1} \phi_{LIN}^\top(\X_1) \boldsymbol{\epsilon}.
    \end{split}
\end{equation*}

Therefore,
\begin{equation*}
    \begin{split}
        &\| \beta_{LIN}^{(t)} -  \beta_{LIN}^* \|\\
        \leq& \left\| \beta_{LIN}^* \right\| \left\| [I - \frac{\lambda}{n} \X_1^\top \X_i]^t \right\|  \\
        &+  \left\| \left[ I - \left[I - \frac{\lambda}{n} \phi_{LIN}^\top(\X_1) \phi_{LIN}(\X_1)\right]^t \right] \left[\phi_{LIN}^\top(\X_1) \phi_{LIN}(\X_1)\right]^{-1} \phi_{LIN}^\top(\X_1) \boldsymbol{\epsilon} \right\|.
    \end{split}
\end{equation*}

Provided that  $\log(1/\delta_1) \leq n/c $, we know from Lemma~\ref{prop: concentrationForXX} that with probability at least $1-\delta_1$, 
\begin{equation*}
\frac{1}{d} \lesssim \frac{1}{n} \sigma_{min}\left(\phi_{LIN}^\top(\X_1) \phi_{LIN}(\X_1)\right) \leq \frac{1}{n} \sigma_{max}\left(\phi_{LIN}^\top(\X_1) \phi_{LIN}(\X_1)\right) \lesssim  \frac{1}{d},
\end{equation*}
Therefore, given that $\lambda < 1/\sigma_n(X^\top X) $ and $t \gg d$, 
\begin{equation*}
    \begin{split}
        &\left\|[ I - \frac{\lambda}{n} \phi_{LIN}^\top(\X_1) \phi_{LIN}(\X_1)]^t \right\| \leq c_2^{t/d} \lesssim O(1/\sqrt{n}),
    \end{split}
\end{equation*}
where $0 < c_2 < 1$ is a given constant.

On the other hand, since $\epsilon_i$ is mean zero, independent, and $\sigma_y$-subGaussian with variance $\sigma^2$. Therefore, following Theorem~{6.3.1} in \citet{vershynin_2018}, with probability at least $1-\delta_2$ (the probability is over $\boldsymbol{\epsilon}$ given $\X_1$),
\begin{equation*}
    \| H \boldsymbol{\epsilon}\| / \sigma \leq  \| H\|_F   + c_3 \sigma_{\x}^2 \| H\|  \sqrt{\log(1/\delta_3)},
\end{equation*}
where $H = \left[ I - \left[I - \frac{\lambda}{n} \phi_{LIN}^\top(\X_1) \phi_{LIN}(\X_1)\right]^t \right] \left[\phi_{LIN}^\top(\X_1) \phi_{LIN}(\X_1)\right]^{-1} \phi_{LIN}^\top(\X_1)$.
Note that since $H \preceq \left[\phi_{LIN}^\top(\X_1) \phi_{LIN}(\X_1)\right]^{-1} \phi_{LIN}^\top(\X_1)$, the eigenvalues of $H$ satisfies
\begin{equation*}
\begin{split}
    \| H\|_F^2 = \sum_{i=1}^d \sigma_i(H)^2 \leq \sum_{i=1}^d \sigma_i(\left[\phi_{LIN}^\top(\X_1) \phi_{LIN}(\X_1)\right]^{-1} \phi_{LIN}^\top(\X_1))^2 \leq \frac{d^2}{n},  \\
    \| H\|_2^2 = \max_i \sigma_i(H)^2 \leq \max_i \sigma_i(\left[\phi_{LIN}^\top(\X_1) \phi_{LIN}(\X_1)\right]^{-1} \phi_{LIN}^\top(\X_1))^2 \leq  \frac{d}{n}.
\end{split}
\end{equation*}
Therefore, with probability at least $1-\delta_2$ with $\log(1/\delta_3) \lesssim \sqrt{d}$,
\begin{equation*}
    \| H\boldsymbol{\epsilon} \| \lesssim  \sqrt{d} \frac{\sqrt{d} + \sigma_{\x}^2 \sqrt{\log(1/\delta_3)}}{\sqrt{n}} \lesssim \frac{d}{\sqrt{n}}.
\end{equation*}

In summary, we have that with high probability
\begin{equation*}
\| \beta_{LIN}^{(t)} -  \beta_{LIN}^* \| \lesssim \| \beta_{LIN}^* \|^2 c_2^{t/d} + \frac{d}{\sqrt{n}} \lesssim \frac{d}{\sqrt{n}}.
\end{equation*}

Therefore, due to the fact that $\| \X\| = O(\sqrt{n} / \sqrt{d})$ and $\| \Z\| = O(\sqrt{n} / \sqrt{d})$, we have that with high probability,
\begin{equation*}
\begin{split}
    &\frac{1}{\sqrt{n}} \| \hat{\Y}_{LIN}(\X)  - \Y(\X) \|_2\\ \leq& 
\frac{1}{\sqrt{n}} \| \hat{\Y}_{LIN}(\X)  - \hat{\Y}_{LIN}^*(\X) \|_2 + \| \hat{\Y}_{LIN}^*(\X)  - \Y(\X) \|_2 \\
\lesssim& \sqrt{\frac{d}{n}} + \epsilon,\\
            &\frac{1}{\sqrt{n}} \| \hat{\Y}_{LIN}(\Z)  - \hat{\Y}_{LIN}^*(\Z) \|_2\lesssim \sqrt{\frac{d}{n}}\\.
\end{split}
\end{equation*}

\end{proof}

\clearpage
\begin{proposition}[Trajectory Analysis]
\label{prop: Trajectory Analysis}
For overparameterized linear regression ($n < p$) with gradient descent starting from $\theta^{(0)} = 0$, we have 
\begin{equation*}
    \theta^{(t)} = \X^\top \left[I -  \left[I - \X \X^\top \right]^t\right] \left[\X \X^\top\right]^{-1} Y.
\end{equation*}

For underparameterized linear regression ($n > p$) with gradient descent starting from $\theta^{(0)} = 0$, we have
\begin{equation*}
    \theta^{(t)} = \X^\top \left[I -  \left[I - \X \X^\top \right]^t\right] \X \left[\X^\top \X\right]^{-2} \X^\top Y.
\end{equation*}
\end{proposition}

\begin{proposition}[Bounding the kernels, from~\citet{DBLP:conf/nips/HuXAP20}, Proposition~{D.2}]
\label{prop: bounding train kernels}
Under the settings in Theorem~\ref{thm: mainTheorem}, 
the following inequality holds with high probability
$$\left\| \phi_{NTK}(X) \phi_{NTK}^\top(X)  - \phi_{LIN}(X) \phi_{LIN}^\top(X)  \right\| \lesssim \frac{n}{d^{1+\alpha}},$$
where the probability is taken over random initialization $\W(0)$ and the training data $\X$.
\end{proposition}

\begin{proposition}[Bounding difference from matrix function, from Theorem~11.4 in \citet{aleksandrov2011estimates}]
\label{prop: matrix function}
For matrix function (function over eigenvalues) with Lipschitz constant $L$, we have that for any matrix $A, B  \in \bR^{n \times n}$, there exists a constant $C$ such that 
$$\| f(A) - f(B) \| \leq C L \log(n) \| A - B\|.$$
\end{proposition}

\begin{proposition}[Bounding the cross kernels]
\label{prop: bounding cross kernels}
Under the settings in Theorem~\ref{thm: mainTheorem}, 
the following inequality holds with high probability
$$\left\| \phi_{NTK}(Z) \phi_{NTK}^\top(X)  - \phi_{LIN}(Z) \phi_{LIN}^\top(X)  \right\| \lesssim \frac{n}{d^{1+\alpha}},$$
where the probability is taken over random initialization $\W(0)$ and the training data $\X$.
\end{proposition}

\begin{proof}
The proof is inspired by~\citet{DBLP:conf/nips/HuXAP20}.
We generalize the results of Theorem~\ref{prop: bounding train kernels} to the cross regimes. 
The proof is divided into three steps. 

\textbf{Step One.}
We firstly show that 
$$\left\| \phi_{NTK}(Z) \phi_{NTK}^\top(X)  - \bE_{\W}\phi_{NTK}(Z) \phi_{NTK}^\top(X) \right\| \lesssim \frac{n}{d^{1+\alpha}},$$
This is due to Bernstein inequality.
Due to the symmetric initialization, we first consider half number of the neurons which are guaranteed to be independent.
For the $r$-th neuron, we denote 
$$B^{(r)} =\left( \sigma^\prime \left(Z w_r / \sqrt{d} \right) \sigma^\prime \left(X w_r / \sqrt{d}\right) \right) \odot \left( Z X^\top/\sqrt{d} \right).$$
Therefore, $\phi_{NTK}(Z) \phi_{NTK}^\top(X) = \frac{1}{m} \sum_{r=1}^m B^{(r)}$, and all neurons $\{ B^{(r)} \}, r=1, \dots, m/2$ are independent. Besides, we show that $B^{(r)}$ is bounded with high probability.
\begin{equation*}
\begin{split}
    \left\|B^{(r)} \right\| &= \left\| \operatorname{diag}\left(\sigma^\prime\left(\Z w_r /\sqrt{d}\right)\right) \cdot \frac{\Z \X^\top}{d} \cdot \operatorname{diag}\left(\sigma^\prime\left(\X w_r /\sqrt{d}\right)\right) \right\|\\
    &\leq  \left\| \operatorname{diag}\left(\sigma^\prime\left(\Z w_r /\sqrt{d}\right)\right) \right\| \left\| \frac{\X}{\sqrt{d}} \right\|  \left\| \frac{\Z }{\sqrt{d}} \right\| \left\|\operatorname{diag}\left(\sigma^\prime\left(\X w_r /\sqrt{d}\right)\right) \right\| \\
    &\leq O(1) \cdot O\left(\sqrt{\frac{n}{d}}\right) \cdot O\left(\sqrt{\frac{n}{d}}\right) \cdot O(1) \\
    &= O\left(\frac{n}{d}\right),
\end{split}
\end{equation*}
where we use the fact that $\| \Z \| \lesssim n$ and $\| \X \| \lesssim n$ using the assumption that $X$ and $\Z$ are diagonalizable.
Therefore, we derive that 
\begin{equation*}
    \begin{split}
        &\left\| B^{(r)} \right\| \leq O(n/d) \\
        &\left\|\bE B^{(r)} \right\| \leq O(n/d) \\
        &\left\|B^{(r)} - \bE B^{(r)} \right\| \leq O(n/d) \\
        &\left\| \sum_{r=1}^{m/2} \bE[(B^{(r)} - \bE B^{(r)})^2] \right\| \leq \sum_{r=1}^{m/2} \left\|\bE(B^{(r)} - \bE B^{(r)})^2 \right\| \leq O(m n^2/d^2)
    \end{split}
\end{equation*}

By Matrix Bernstein inequality, we have that 
\begin{equation*}
    \bP\left[ \left\| \sum_{i=1}^{m/2} B^{(r)} - \bE B^{(r)} \right\| \geq \frac{m}{2} \frac{n}{d^{1+\alpha}} \right] \ll 1,
\end{equation*}
where we use $m = \Omega(d^{1+\alpha})$ and $n = d^{O(1)}$.

Since $\phi_{NTK}(Z) \phi_{NTK}^\top(X) = \frac{1}{m} \sum_{r=1}^m B^{(r)}$
We rewrite the above inequality as follows: with high probability,
\begin{equation*}
\begin{split}
    &\left\| \phi_{NTK}(Z) \phi_{NTK}^\top(X)  - \bE_{\W}\phi_{NTK}(Z) \phi_{NTK}^\top(X) \right\| \\
    =& \left\| \frac{1}{m} \sum_{r=1}^m [B^{(r)} - \bE B^{(r)}] \right\|\\
    \leq & 2 \left\| \frac{1}{m/2} \sum_{r=1}^{m/2} [B^{(r)} - \bE B^{(r)}] \right\|\\
    \leq & \frac{1}{m} O(m n/d^{1+\alpha}) \\
    \lesssim & \frac{n}{d^{1+\alpha}}.
\end{split}
\end{equation*}
$$$$

\textbf{Step Two.} We secondly show that 
$$\left\| \bE_{\W}\phi_{NTK}(Z) \phi_{NTK}^\top(X) - \phi_{LIN}(Z) \phi_{LIN}^\top(X)  \right\| \lesssim \frac{n}{d^{1+\alpha}}.$$

By defining $\Phi(a, b, c) = \bE_{(u_1, u_2) \sim \cN(0, \Lambda) }[\sigma^\prime(u_1) \sigma^\prime(u_2)]$, where $\Lambda = 
\left( \begin{array}{cc}
    a & c \\
    c & b
\end{array} \right)
$, we rewrite that,
\begin{equation*}
\begin{split}
    \left[\bE_{\W}\phi_{NTK}(Z) \phi_{NTK}^\top(X)\right]_{ij} &= \frac{1}{d} z_i^\top x_j \cdot \bE_{w \sim \cN(0, I)} \left[\sigma^\prime (w^\top z_i /\sqrt{d}) \sigma^\prime (w^\top x_i /\sqrt{d})^\top \right]\\
    &= \frac{1}{d} z_i^\top x_j \Phi\left( \frac{\| z_i\|^2}{d}, \frac{\| x_i\|^2}{d}, \frac{z_i^\top x_j}{d} \right).
\end{split}
\end{equation*}
Due to the assumptions on the data distribution and $n \gg d$, we have that
$\frac{\|x_i\|^2}{d} = 1 \pm \tilde{O}(\frac{1}{\sqrt{d}})$, $\frac{\|z_j\|^2}{d} = 1 \pm \tilde{O}(\frac{1}{\sqrt{d}})$,
and $\frac{z_j^\top x_i}{d} =  \pm \tilde{O}(\frac{1}{\sqrt{d}})$.
We apply Taylor expansion of $\Phi$ around (1, 1, 0), that is 
\begin{equation*}
    \begin{split}
        &\Phi\left( \frac{\| z_i\|^2}{d}, \frac{\| x_i\|^2}{d}, \frac{z_i^\top x_j}{d} \right) \\
        =& \Phi(1, 1, 0) + c_1 \left(\frac{\| z_i\|^2}{d} - 1\right) +c_2 \left(\frac{\| x_i\|^2}{d} - 1\right) + c_3 \left(\frac{z_i^\top x_j}{d} \right) \pm \tilde{O}\left(\frac{1}{d}\right);
    \end{split}
\end{equation*}
Similar to the proof in \citet{DBLP:conf/nips/HuXAP20}, we have
\begin{equation*}
    \left\|\bE_{\W}\phi_{NTK}(Z) \phi_{NTK}^\top(X)- \zeta^2 \frac{Z X^\top}{d} - c_3 \frac{Tr[\Sigma^2]}{d^2} \boldsymbol{1}\boldsymbol{1}^\top \right\| \leq \tilde{O}(\frac{n}{d^{1.25}}).
\end{equation*}
which is equivalent to 
\begin{equation*}
\left\| \bE_{\W}\phi_{NTK}(Z) \phi_{NTK}^\top(X) - \phi_{LIN}(Z) \phi_{LIN}^\top(X)  \right\| \lesssim \frac{n}{d^{1+\alpha}},
\end{equation*}
where $0<\alpha<1/4$.

Combining the two steps, we have that 
$$\left\| \phi_{NTK}(Z) \phi_{NTK}^\top(X)  - \phi_{LIN}(Z) \phi_{LIN}^\top(X)  \right\| \lesssim \frac{n}{d^{1+\alpha}}.$$
\end{proof}

\begin{proposition}
\label{prop: ex function Lip}
Denote function $e(x) = \frac{1 - (1- x)^t}{ x}$, its Lipschitz constant satisfies $L_e \lesssim t^2$.
\end{proposition}

\begin{proof}[Proof of Proposition~\ref{prop: ex function Lip}]
Note that 
\begin{equation*}
    \begin{split}
        e^\prime(x) &= \frac{tx (1-x)^{t-1} + (1-x)^t - 1}{x^2} \\
        &= \frac{t(1-x)^{t-1}}{x} + \frac{(1-x)^t - 1}{x^2} \\
    \end{split}
\end{equation*}
is decreasing with $x$, where $x \in (0, 1)$ and $t \gg 3$.
Therefore, 
\begin{equation*}
    \begin{split}
        e^\prime(x) &\leq e^\prime(0) = -\frac{t(t-1)}{2} <0.
    \end{split}
\end{equation*}

Therefore, we consider $|e^\prime(x)| = - e^\prime(x)$. Note that by inequality $(1-x)^{t} \geq 1 - tx$, we have
\begin{equation*}
    \begin{split}
        |e^\prime(x)| &= - e^\prime(x) \\
        &= \frac{-tx (1-x)^{t-1} - (1-x)^t + 1}{x^2} \\
        &\leq \frac{-tx (1-({t-1})x) - (1-tx) + 1}{x^2} \\
        &= t(t-1) \\
        &\lesssim t^2.
    \end{split}
\end{equation*}
Therefore, $L_e \lesssim t^2$.

\end{proof}

\begin{proposition}
\label{prop: concentrationForXX}
Assume that $\Sigma_{\x}^{-1/2} \x_i$ is mean zero, independent, and $\sigma_{\x}$-subGaussian ($\| \cdot \|_{\psi_2} \leq \sigma_{\x}$), 
under the assumption $n \gg d$,
then the maximal and minimal eigenvalue of $\X^\top \X$ satisfy: with probability at least $1 - \delta$:
\begin{equation*}
\|\Sigma_{\x} \|_{min} (n - C_1 \sigma_{\x}^2 \log(1/\delta) )\lesssim \sigma_n(X^\top X) \leq \sigma_1(X^\top X) \lesssim  \|\Sigma_{\x} \|_{max}( n + C_1 \sigma_{\x}^2 \log(1/\delta)),
\end{equation*}
where $\| \Sigma_{\x} \|_{min}$, $\| \Sigma_{\x} \|_{max}$ represent the minimal and maximal eigenvalues of $\Sigma_{\x}$ and $C_1$ is a universal constant independent of $n$ and $d$. 
\end{proposition}

\begin{proof}
Denote $A \triangleq \X\Sigma_{\x}^{-1/2}$ where $\X \in \bR^{n\times d}$ is the corresponding design matrix.
Note that each row of matrix $A$ is $A_i = \x_i^\top \Sigma_{\x}^{-1/2} $, which is independent and $\Sigma_{\x}$-subGaussian, and $\bE A_i = \boldsymbol{0}$.
Therefore, the maximum $\sigma_1(A)$ and minimal $\sigma_n(A)$ eigenvalue of $A$ satisfy the following inequality with probability at least $1- 2\exp(-t^2)$
\begin{equation*}
    \sqrt{n} - c_1 \Sigma_{\x} (\sqrt{d} + t) \leq \sigma_n(A) \leq \sigma_1(A) \leq \sqrt{n} + c_1 \Sigma_{\x} (\sqrt{d} + t),
\end{equation*}
where $c_1$ is a constant.

Due to the assumption $n \gg d$, it holds that with probability at least $1-\delta$, 
\begin{equation*}
 \sqrt{n} - c_1 \Sigma_{\x} \sqrt{\log(1/\delta)} \lesssim \sigma_n(A) \leq \sigma_1(A) \lesssim \sqrt{n} + c_1 \Sigma_{\x} \sqrt{\log(1/\delta)}.   
\end{equation*}

Since $\sigma_i(A^\top A) = \sigma_i(A)^2$, we have
\begin{equation*}
n - c_1^2 \Sigma_{\x}^2 \log(1/\delta) \lesssim \sigma_n(A^\top A) \leq \sigma_1(A^\top A) \lesssim n + c_1^2 \Sigma_{\x}^2 \log(1/\delta).   
\end{equation*}

Note that $A^\top A = \Sigma_{\x}^{-1/2} \X^\top \X \Sigma_{\x}^{-1/2}$, we have
\begin{equation*}
\|\Sigma_{\x} \|_{min} (n - c_1^2 \Sigma_{\x}^2 \log(1/\delta) )\lesssim \sigma_n(X^\top X) \leq \sigma_1(X^\top X) \lesssim  \|\Sigma_{\x} \|_{max}( n + c_1^2 \Sigma_{\x}^2 \log(1/\delta)).   
\end{equation*}

\end{proof}